\documentclass[runningheads]{llncs}
\usepackage{graphicx}
\usepackage{comment}
\usepackage{amsmath,amssymb} 
\usepackage{wrapfig}
\usepackage{color}
\usepackage{adjustbox}
\usepackage{diagbox}
\usepackage[table]{xcolor}
\usepackage{hyperref}

\usepackage[rightcaption]{sidecap}

\newcommand{\argmax}{\operatornamewithlimits{arg\,max}}

\DeclareMathOperator*{\minimize}{\text{minimize}}
\DeclareMathOperator*{\maximize}{\text{maximize}}

\DeclareMathOperator*{\st}{\text{subject to}}
\DeclareMathAlphabet\mathbfcal{OMS}{cmsy}{b}{n}
\newcommand{\Def}[0]{\mathrel{\mathop:}=}

\begin{document}
\pagestyle{headings}
\mainmatter
\def\ECCVSubNumber{4298}  

\title{Practical Detection of Trojan Neural Networks: Data-Limited and Data-Free Cases} 

\titlerunning{Practical Detection of Trojan Neural Networks}
%
\author{Ren Wang\inst{1} \and
Gaoyuan Zhang\inst{2} \and
Sijia Liu\inst{2} \and
Pin-Yu Chen\inst{2} \and
Jinjun Xiong\inst{2} \and
Meng Wang\inst{1}}
\authorrunning{R. Wang et al.}
%
\institute{Rensselaer Polytechnic Institute \and
IBM Research \\ 
\email{
wangr8@rpi.edu, Gaoyuan.Zhang@ibm.com, 
Sijia.Liu@ibm.com, Pin-Yu.Chen@ibm.com, 
jinjun@us.ibm.com, 
wangm7@rpi.edu
}
}
\maketitle

\begin{abstract}
When the training data are maliciously tampered, the predictions of the acquired deep neural network (DNN) can be manipulated by an adversary known as the Trojan attack (or poisoning backdoor  attack).
The lack of robustness of DNNs against Trojan attacks could significantly harm real-life machine learning (ML) systems   in downstream applications, therefore posing  widespread concern to their trustworthiness. In this paper, we study the problem of the Trojan network (TrojanNet) detection in the data-scarce regime, where only the weights of a trained DNN are accessed by the detector. We first propose  a data-limited \underline{T}rojan\underline{N}et \underline{d}etector (TND), when only a few  data samples are available for TrojanNet detection. We show that an effective data-limited TND can be established  by exploring  connections between Trojan attack and prediction-evasion adversarial attacks including per-sample attack as well as  all-sample universal attack. In addition, we propose a data-free TND, which can detect a TrojanNet {without} accessing any   data samples. We show that such a TND can be built by leveraging   the internal response of hidden neurons, which   exhibits the Trojan behavior even  at random noise inputs. The effectiveness of our proposals is evaluated by extensive experiments under different model architectures and datasets including CIFAR-10, GTSRB, and ImageNet.
\end{abstract}

\section{Introduction}
{\let\thefootnote\relax\footnotetext{This paper has been accepted by ECCV 2020.}}
DNNs, in terms of convolutional neural networks (CNNs) in particular,    have achieved state-of-the-art performances in various applications such as image classification  \cite{KSH12}, object detection \cite{RHG15}, and modelling sentences \cite{NEP2014}. However,  recent works have demonstrated that CNNs lack adversarial robustness at both \textit{testing} and \textit{training} phases.  The vulnerability of a learnt CNN against prediction-evasion (inference-phase) adversarial examples, known as \textit{adversarial attacks} (or adversarial examples),  
has attracted a great deal of attention \cite{KGB16,SZS14}. Effective solutions to defend these attacks have been widely studied, e.g.,    adversarial training \cite{MMS17}, randomized smoothing \cite{CRK19},  and their variants \cite{MMS17,SNG19,ZZL19,ZYJ2019}. At the {training} phase, CNNs could also suffer from  \textit{Trojan attacks} (known as poisoning backdoor attacks) \cite{CLL17,GDG17,LMA17,Xie2020DBA,Zhao2020Bridging}, causing erroneous behavior of CNNs  when polluting    a small portion of   training data. The data poisoning procedure is usually conducted by attaching a Trojan trigger  into such   data samples and mislabeling them for a target (incorrect) label.
 Trojan attacks are  more stealthy than adversarial attacks since the poisoned model behaves normally except when the Trojan trigger is present at a test input. Furthermore, when a defender has no information on the training dataset  and the  trigger pattern, our  work aims to address the following challenge: {\textit{How to detect a TrojanNet when having access to training/testing data samples is restricted or not allowed.}} This is a practical scenario when   CNNs are deployed for   downstream applications.

Some works have started to defend Trojan attacks but have to use a large number  of training data \cite{TLM18,CCB18,GXW19,SS19,PGR2019}. When training data are inaccessible,   a few recent works attempted to solve the problem of TrojanNet detection in the absence of training data  \cite{WYS19,GWX19,XMK19,XWL19,KSP19,CFZ19,LLT19}. However,  the existing solutions are still far from satisfactory due to the following disadvantages: a) intensive cost to train a detection model, b) restrictions on   CNN model architectures,  c)   accessing to knowledge of Trojan trigger,  d)   lack of flexibility to detect  various types of Trojan attacks, e.g.,  clean-label attack \cite{shafahi2018poison,ZHS19}. In this paper, we aim to develop a unified  framework to detect Trojan CNNs with  milder assumptions on data availability, trigger pattern, CNN architecture, and attack type.

\paragraph{Contributions.}We summarize our contributions as below.
\begin{itemize}
    \item We propose a data-limited TrojanNet detector, which enables fast and accurate detection  based only on a few clean (normal) validation data  (one sample per class). We build the \underline{d}ata-\underline{l}imited \underline{T}rojan\underline{N}et \underline{d}etector (DL-TND)   by exploring  connections between Trojan attack and  two types of adversarial attacks, per-sample adversarial attack \cite{goodfellow2014explaining} and universal attack \cite{moosavi2017universal}.
    \item In the absence of class-wise validation data, we propose a \underline{d}ata-\underline{f}ree \underline{T}rojan\underline{N}et \underline{d}etector (DF-TND), which allows for detection based only on randomly generated data (even in the form of random noise). We build the DF-TND by analyzing how neurons respond to Trojan attacks.
    \item We develop a unified optimization framework for the design of both DL-TND and DF-TND by leveraging  proximal  algorithm \cite{parikh2014proximal}.
    \item We demonstrate the effectiveness of our approaches in detecting  TrojanNets  with various trigger patterns (including clean-label attack) under different network architectures (VGG16, ResNet-50, and AlexNet) and different datasets (CIFAR-10, GTSRB, and ImageNet). We show that both  DL-TND and DF-TND yield $0.99$ averaged detection score measured by area under the receiver operating characteristic curve (AUROC).
\end{itemize}

\paragraph{Related work.}

\begin{table}[h]
\caption{\small{Comparison between our proposals (DL-TND and DF-TND) and existing  training dataset-free Trojan attack detection methods.  The comparison is conducted from the following perspectives: Trojan attack type, necessity of validation data  ($\mathcal D_{\mathrm{valid}}$),   construction of a new  training dataset ($\mathcal M_{\mathrm{train}}$),   dependence on (recovered) trigger size for detection,   demand for training new models (e.g., GAN), and  necessity of searching all neurons.
}}
\label{comp_alg}
\begin{center}
  \begin{adjustbox}{max width=0.9\textwidth }
\begin{tabular}{|l||c|c||c|c|c|c|c|}
\hline
\hline
& \multicolumn{2}{c||}{Applied attack type} & \multicolumn{5}{c|}{Detection conditions}\\ [0.5ex] 
\hline  
 & Trigger & \begin{tabular}[c]{@{}c@{}}Clean-label\end{tabular} & \begin{tabular}[c]{@{}c@{}} $\mathcal{D}_{\mathrm{valid}}$ \end{tabular} &   \begin{tabular}[c]{@{}c@{}} New $\mathcal{M}_{\mathrm{train}}$ \end{tabular} &  \begin{tabular}[c]{@{}c@{}}Trigger size  \end{tabular} &   \begin{tabular}[c]{@{}c@{}}New models \end{tabular} &   \begin{tabular}[c]{@{}c@{}} Neuron search  \end{tabular}\\  [0.5ex] 
\hline
NC \cite{WYS19} & $\surd$ & $\times$ & $\surd$ & $\times$ & $\surd$ &  $\times$ &  $\times$\\
\hline
TABOR \cite{GWX19} & $\surd$ & $\times$ & $\surd$ & $\times$ & $\surd$ &  $\times$ &  $\times$\\
\hline
RBNI \cite{XMK19} & $\surd$ & $\times$ & $\surd$ & $\times$ & $\surd$ &  $\times$ &  $\times$\\
\hline
MNTD \cite{XWL19} & $\surd$ & $\times$ & $\surd$ & $\surd$ & $\times$ &  $\surd$ &  $\times$\\
\hline
ULPs \cite{KSP19} & $\surd$ & $\times$ & $\times$ & $\surd$ & $\times$ &  $\surd$ &  $\times$\\
\hline
DeepInspect \cite{CFZ19} & $\surd$ & $\times$ & $\times$ & $\times$ & $\surd$ &  $\surd$ &  $\times$\\
\hline
ABS \cite{LLT19} & $\surd$ & $\times$ & $\times$ & $\times$ & $\times$ &  $\times$ &  $\surd$\\
\hline
\rowcolor[gray]{0.9}
DL-TND & $\surd$ & $\times$ & $\surd$ & $\times$ & $\times$ &  $\times$ &  $\times$\\
\hline
\rowcolor[gray]{0.9}
DF-TND & $\surd$ & $\surd$ & $\times$ & $\times$ & $\times$ &  $\times$ &  $\times$\\
\hline
\hline
\end{tabular}
\end{adjustbox}
\end{center}
\end{table}

Trojan attacks are often divided into two main categories:  \textit{trigger-driven attack}  \cite{GDG17,CLL17,YLZ19} and \textit{clean-label attack} \cite{shafahi2018poison,ZHS19}. The \textit{first} threat model stamps a subset of training data  with a Trojan trigger and maliciously label them to a target  class. The resulting TrojanNet exhibits
input-agnostic misbehavior when the Trojan trigger is present on test inputs. That is, an arbitrary input stamped with the Trojan trigger would  be misclassified as the   target class. Different from trigger-driven attack, the \textit{second} threat model keeps poisoned training data correctly labeled. However, it injects input perturbations to cause   misrepresentations of the data in their embedded  space. Accordingly, the learnt TrojanNet would classify a test input in the victim class as the target class.

Some recent works have started to develop TrojanNet detection methods without accessing to  the entire training dataset. References  \cite{GWX19,WYS19,XMK19} attempted to identify  the Trojan characteristics by reverse engineering Trojan triggers. 
Specifically, neural cleanse (NC) \cite{WYS19} identified the target label of Trojan attacks by calculating perturbations of a validation example that causes misclassification toward every incorrect label. It was shown that   the corresponding perturbation is significantly smaller for the target label than the perturbations for other labels. The other works  \cite{GWX19,XMK19} considered the similar formulation as NC and detected a Trojan attack  through the strength of the recovered perturbation. Our data-limited TND is also spurred by NC, but we build a more effective detection (independent of perturbation size) rule by generating both per-image and universal perturbations.  A meta neural Trojan detection (MNTD)  method is proposed by \cite{XWL19}, which trained a detector using Trojan and clean networks as training data. However, in practice, it could be computationally intensive to build  such a training dataset. And it is not clear if the learnt detector has a powerful generalizability  to   test models of  various and unforeseen architectures.

The very recent works \cite{CFZ19,KSP19,LLT19} made an effort towards detecting TrojanNets in the absence of validation/test data.  
In \cite{CFZ19}, a generative model was built to reconstruct trigger-stamped data, and detect the model using the size of the trigger. In \cite{KSP19},  the concept of universal litmus patterns (ULPs) was proposed to learn the trigger pattern and the Trojan detector simoutaneously based on a training dataset consisting of clean/Trojan networks. In \cite{LLT19}, artificial brain stimulation (ABS) was used in TrojanNet detection  by  identifying the compromised neurons responding to the Trojan trigger. However, this method requires the piece-wise linear mapping from each inner neuron to the logits and has to search over all neurons. Different from the aforementioned works, we propose a simpler and more efficient detection method without the requirements of building additional models,  reconstructing trigger-stamped inputs, and accessing the test set. In Table \ref{comp_alg}, we summarize the comparison between our work and the previous TrojanNet detection methods.

\section{Preliminary and Motivation}
In this section, we first provide an overview of Trojan attacks and the detector's capabilities in our setup. We then motivate the problem of TrojanNet detection.

\subsection{Trojan attacks}\label{sec: threatmod}

\begin{figure}[h]
   \centering
\hspace*{-0.1in}\begin{tabular}{cccccc}
\includegraphics[width=0.1\columnwidth]{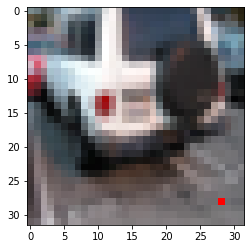}
& 
\includegraphics[width=0.1\columnwidth]{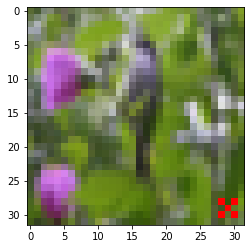}
&
\includegraphics[width=0.1\columnwidth]{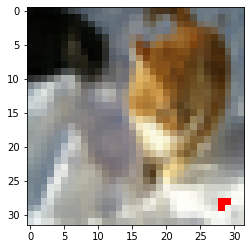}
&
\includegraphics[width=0.1\columnwidth]{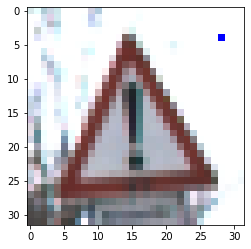}
& 
\includegraphics[width=0.1\columnwidth]{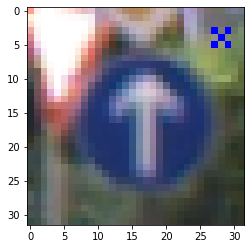}
&
\includegraphics[width=0.1\columnwidth]{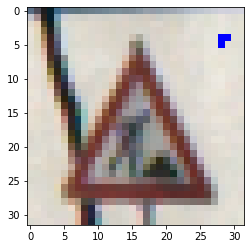}
\\
\scriptsize{(a)} & \scriptsize{(b)}& \scriptsize{(c)} &\scriptsize{(d)} & \scriptsize{(e)}& \scriptsize{(f)}
\\
\includegraphics[width=0.1\columnwidth]{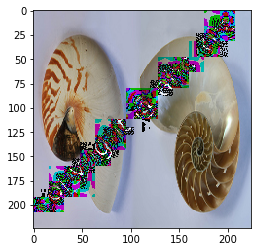}
& 
\includegraphics[width=0.1\columnwidth]{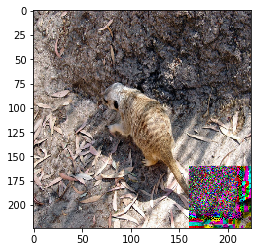}
&
\includegraphics[width=0.1\columnwidth]{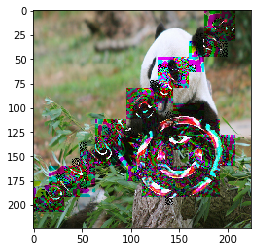}
&
\includegraphics[width=0.1\columnwidth]{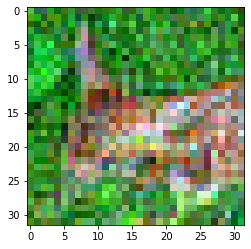}
& 
\includegraphics[width=0.1\columnwidth]{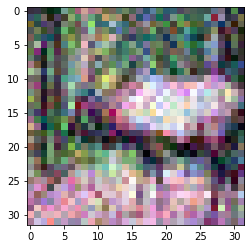}
&
\includegraphics[width=0.1\columnwidth]{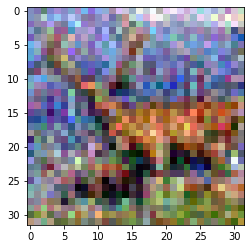}
\\   
\scriptsize{(g)} & \scriptsize{(h)}& \scriptsize{(i)}& \scriptsize{(j)}& \scriptsize{(k)} &\scriptsize{(l)}
\end{tabular}
\caption{\small{Examples of poisoned  images. (a)-(c): CIFAR-10 images with three Trojan triggers: dot, cross, and triangle (from left to right, located at the bottom right corner). (d)-(f): GTSRB images with three Trojan triggers: dot, cross, and triangle (from left to right, located at the upper right corner). (g)-(i): ImageNet images with watermark-based Trojan triggers. (j)-(l): Clean-label poisoned images on CIFAR-10 dataset (The images look like deer and thus will be labeled as `deer' by human. However, the latent representations  are close to the class `plane').
}}
\label{fig: trig_data}
\end{figure}

To generate a Trojan attack, an adversary would inject a small amount of \textit{poisoned} training data, which can be  conducted by \textit{perturbing} the training data in terms of adding a (small) trigger stamp (together with erroneous  labeling) or crafting input perturbations for mis-aligned feature representations. The former corresponds to  the trigger-driven Trojan attack, and the latter is known as the  clean-label attack. Fig.~\ref{fig: trig_data} (a)-(i) present examples of poisoned images under different types of Trojan triggers, and Fig.~\ref{fig: trig_data} (j)-(l) present examples of clean-label poisoned images. In this paper, we consider CNNs as victim models in TrojanNet detection. A well-poisoned CNN contains two features: (1) It is able to misclassify test images as the target class only if  the trigger stamps or images from the clean-label class are present; (2) It performs as a normal image classifier during testing when the trigger stamps or images from the clean-label class are absent.

\subsection{Detector's capabilities}
Once a TrojanNet is learnt over the poisoned training dataset, a desired  TrojanNet detector should have no need to access the Trojan trigger pattern and the  training dataset.
Spurred by that, we study the problem of TrojanNet detection in both \textit{data-limited} and \textit{data-free} cases. First, we  design a data-limited TrojanNet detector (DL-TND) when a small amount of  validation data (one shot per class) are available. Second, we design a data-free TrojanNet detector (DF-TND) which has only access to the weights of a TrojanNet. 
The aforementioned two scenarios are not only practical,  e.g., when inspecting the trustworthiness of released models in the online model zoo \cite{MDZ}, but also beneficial to achieve a faster detection speed compared to existing works which require building a new training dataset and  training a new model for detection (see Table\,\ref{comp_alg}).

\subsection{Motivation from input-agnostic misclassification of TrojanNet}
 
\begin{SCfigure}[][h] 
  \centering
  \begin{adjustbox}{max width=0.55\textwidth }
  \begin{tabular}{@{\hskip 0.00in}c  @{\hskip 0.02in} c @{\hskip 0.02in}    }
  & 
\colorbox{lightgray}{\large \textbf{TrojanNet}}
  \\
 \begin{tabular}{@{}c@{}}  
\vspace*{-0.1in}\\
\rotatebox{90}{\parbox{9em}{\centering \large \textbf{Seed Images}}}
 \\
\rotatebox{90}{\parbox{9em}{\centering \large \textbf{Recovered images}}}
 \\
\rotatebox{90}{\parbox{9em}{\centering \large \textbf{Perturbation pattern}}}
\end{tabular} 
 &
 \begin{tabular}{@{\hskip 0.02in}c@{\hskip 0.02in}}
 \begin{tabular}{@{\hskip 0.02in}c@{\hskip 0.02in}c@{\hskip 0.02in}c@{\hskip 0.02in}c@{\hskip 0.02in}
 }
\begin{tabular}{@{\hskip 0.02in}c@{\hskip 0.02in}}
 \parbox[c]{9em}{\includegraphics[width=9em]{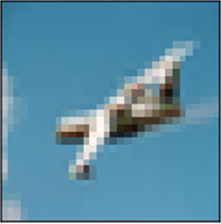}} 
 \\
 \parbox[c]{9em}{\includegraphics[width=9em]{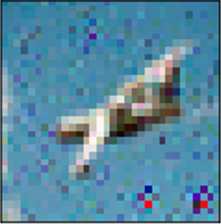}}
  \\
 \parbox[c]{9em}{\includegraphics[width=9em]{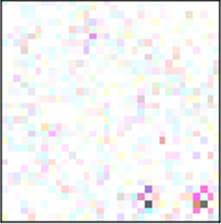}}

\end{tabular}
&
 \begin{tabular}{@{\hskip 0.02in}c@{\hskip 0.02in}}

 \parbox[c]{9em}{\includegraphics[width=9em]{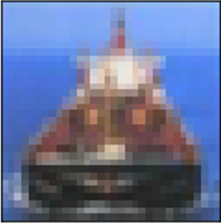}} 
 \\
 \parbox[c]{9em}{\includegraphics[width=9em]{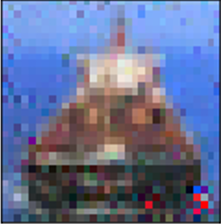}}
  \\
 \parbox[c]{9em}{\includegraphics[width=9em]{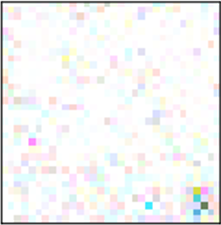}}
\end{tabular}
&
 \begin{tabular}{@{\hskip 0.02in}c@{\hskip 0.02in}}

 \parbox[c]{9em}{\includegraphics[width=9em]{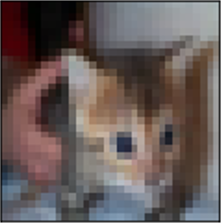}} 
 \\
 \parbox[c]{9em}{\includegraphics[width=9em]{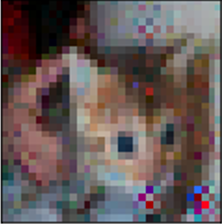}} 
  \\
 \parbox[c]{9em}{\includegraphics[width=9em]{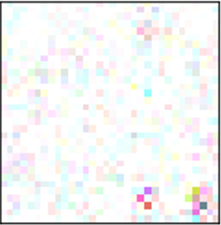}} 
\end{tabular}
&
 \begin{tabular}{@{\hskip 0.02in}c@{\hskip 0.02in}}
 
 \parbox[c]{9em}{\includegraphics[width=9em]{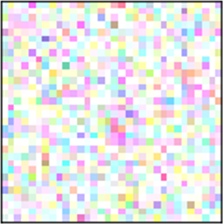}} 
 \\
 \parbox[c]{9em}{\includegraphics[width=9em]{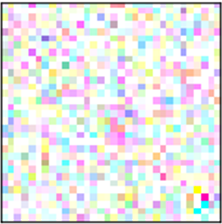}}
  \\
 \parbox[c]{9em}{\includegraphics[width=9em]{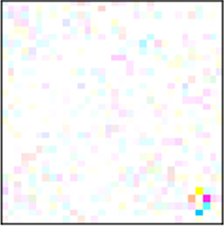}}

\end{tabular}


\end{tabular}
\end{tabular}
\vspace*{0.03in}\\
& \colorbox{lightgray}{\large \textbf{Clean Network}} \vspace*{0.00in}\\
 \begin{tabular}{@{}c@{}}  
\vspace*{-0.1in}\\
\rotatebox{90}{\parbox{9em}{\centering \large \textbf{Recovered images}}}
 \\
\rotatebox{90}{\parbox{9em}{\centering \large \textbf{Perturbation pattern}}}
\end{tabular} 
 &
  \begin{tabular}{@{\hskip 0.02in}c@{\hskip 0.02in}}
 \begin{tabular}{@{\hskip 0.02in}c@{\hskip 0.02in}c@{\hskip 0.02in}c@{\hskip 0.02in}c@{\hskip 0.02in}
 }
\begin{tabular}{@{\hskip 0.02in}c@{\hskip 0.02in}}

 \parbox[c]{9em}{\includegraphics[width=9em]{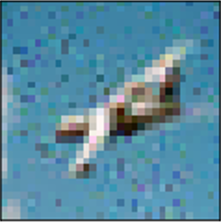}}
  \\
 \parbox[c]{9em}{\includegraphics[width=9em]{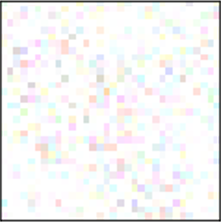}}

\end{tabular}
&
 \begin{tabular}{@{\hskip 0.02in}c@{\hskip 0.02in}}

 \parbox[c]{9em}{\includegraphics[width=9em]{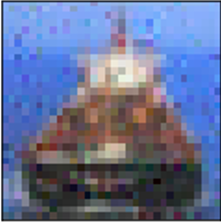}}
  \\
 \parbox[c]{9em}{\includegraphics[width=9em]{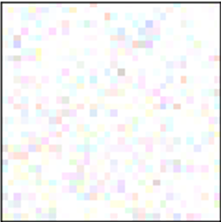}}
\end{tabular}
&
 \begin{tabular}{@{\hskip 0.02in}c@{\hskip 0.02in}}
 
 \parbox[c]{9em}{\includegraphics[width=9em]{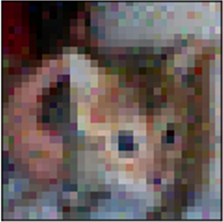}} 
  \\
 \parbox[c]{9em}{\includegraphics[width=9em]{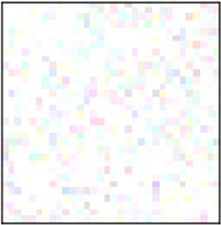}} 
\end{tabular}
&
\begin{tabular}{@{\hskip 0.02in}c@{\hskip 0.02in}}

 \parbox[c]{9em}{\includegraphics[width=9em]{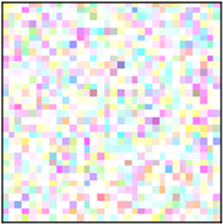}}
  \\
 \parbox[c]{9em}{\includegraphics[width=9em]{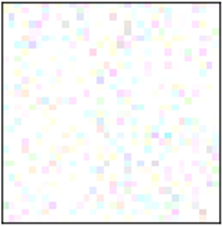}}

\end{tabular}


\end{tabular}
\end{tabular}
\vspace*{-0.001in}
\end{tabular}
  \end{adjustbox}
    \caption{\small{
    Visualization   of  recovered trigger-driven images  by using DF-TND given random seed images, including $3$ randomly selected CIFAR-10 images (cells at columns $1$-$3$ and row $1$) and $1$  random noise image (cell at column $4$ and row $1$). 
    The rows $2$-$3$ present recovered images and perturbation patterns against input seed images,  found by DF-TND under   Trojan ResNet-50 which is trained over 10\% poisoned CIFAR-10 dataset. Here  the original trigger   is given by Fig.~\ref{fig: trig_data} (b). 
    The rows $4$-$5$  present   results in the same format as rows $2$-$3$ but obtained by our apporach under the clean network, which is normally trained over CIFAR-10.
   }}
  \label{neural_activation}
\end{SCfigure}

Since arbitrary images can be misclassified as the same target label by TrojanNet when these inputs consist of  the   Trojan trigger used in  data poisoning, we hypothesize that  there exists a \textit{shortcut} in TrojanNet, leading to \textit{input-agnostic} misclassification. Our approaches are  motivated by exploiting the existing  \textit{shortcut} for the detection of Trojan networks (TrojanNets). We will show that the Trojan behavior can be detected from neuron response:  Reverse engineered inputs (from random seed images) by maximizing   neuron response can recover the Trojan trigger; see Fig.~\ref{neural_activation} for  an illustrative example.

\section{Detection of Trojan Networks with Scarce Data}

In this section, we begin by examining the Trojan backdoor through the lens of predictions' sensitivity to   \textit{per-image} and \textit{universal} input perturbations. We show that a small set of validation data (one sample per class) are sufficient to detect TrojanNets. Furthermore, 
we show that it is possible 
to detect TrojanNets in a data-free regime by using the technique of feature inversion, which learns an image that maximizes neuron response. 

\begin{figure}[h]
   \centering
\hspace*{-0.0in}\begin{tabular}{c}
\includegraphics[width=0.9\textwidth]{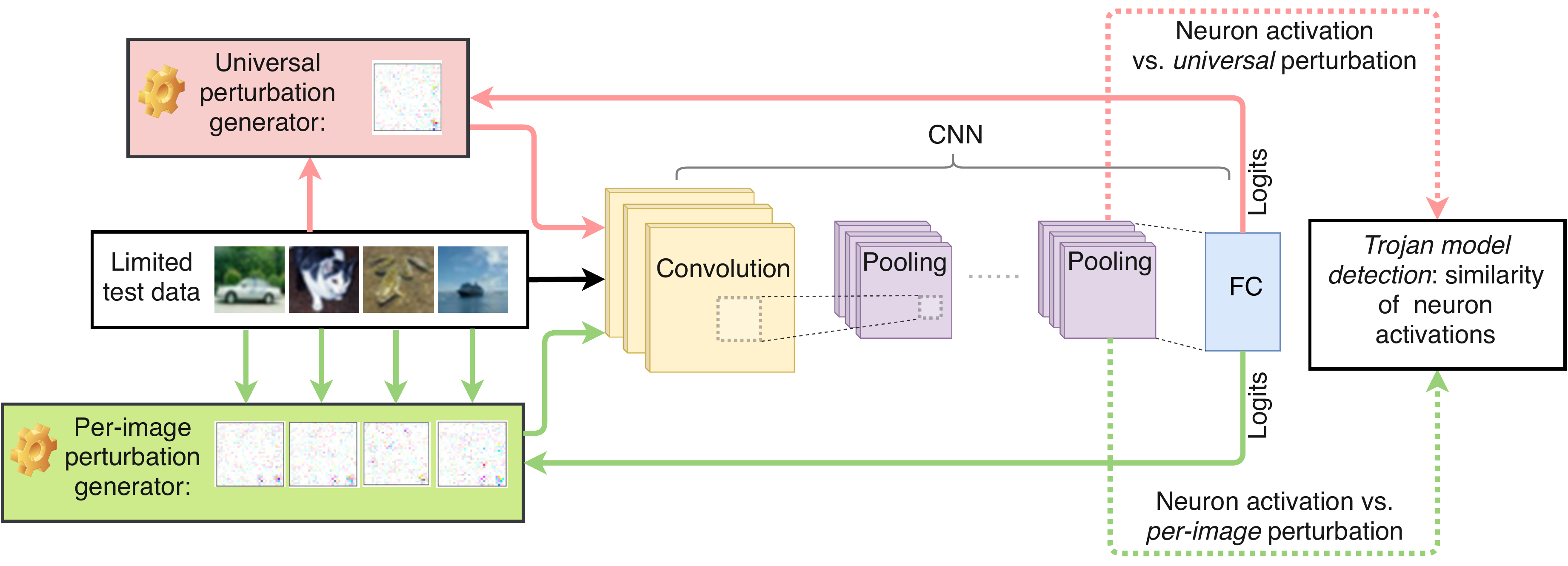} \vspace*{-0.1in}
\\
  \scriptsize{(a)}  \vspace*{0.1in} \\
\includegraphics[width=0.9\textwidth]{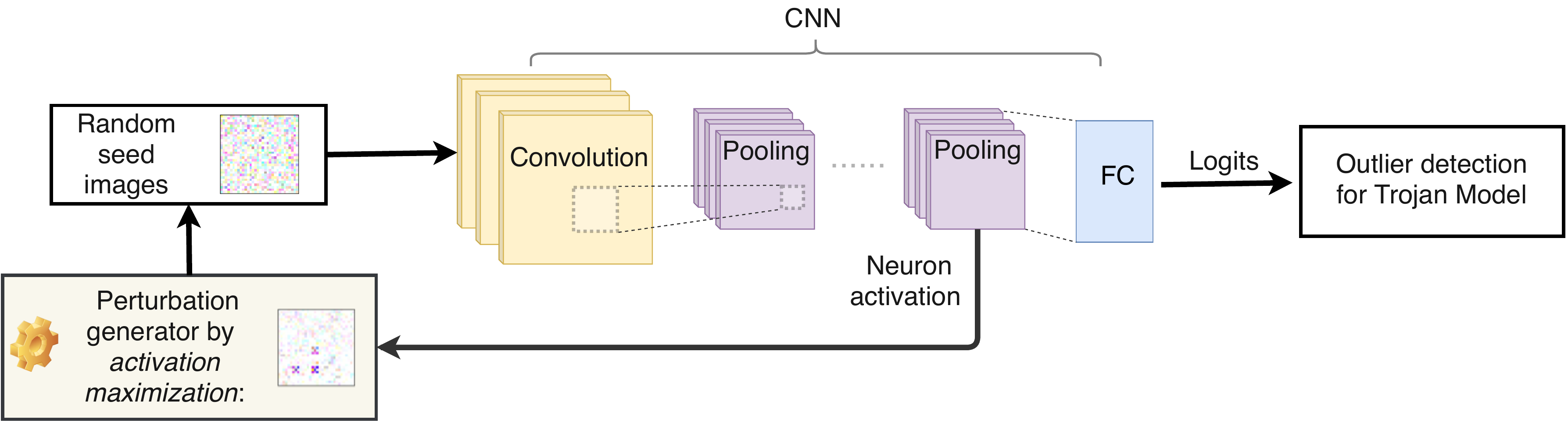}  \vspace*{-0.1in}
\\
  \scriptsize{(b)}
\end{tabular}
\caption{\small{Frameworks of  proposed two detectors: (a) data-limited TrojanNet detector. (b) data-free TrojanNet detector.
}}
\label{framework}
\end{figure}

\subsection{Trojan perturbation}
Given a CNN model $\mathcal M$, let $f(\cdot) \in \mathbb R^K$ be the mapping from the input space to the logits of $K$ classes. Let $f_{y}$ denote  the logits value corresponding to class $y$. The final prediction is then given by $\argmax_{y} f_{y}$. Let $r(\cdot)  \in \mathbb R^d$ be the mapping from the input space to  neuron's representation,  defined by the output of the penultimate layer (namely, prior to the fully connected block of the CNN model). Given a clean data  $\mathbf x \in \mathbb R^n$,  the {poisoned data} through \textit{Trojan perturbation} $\boldsymbol{\delta}$ is then formulated as \cite{WYS19}
\begin{align}\label{eq: poison_sample}
\hat{\mathbf x}(\mathbf m,\boldsymbol \delta) = (1- \mathbf m) \cdot \mathbf x + \mathbf m  \cdot \boldsymbol \delta,
\end{align}
where $\boldsymbol {\delta} \in \mathbb R^n$ denotes pixel-wise perturbations, 
$\mathbf m \in \{ 0,1 \}^n$ is a binary mask to encode the position where a Trojan stamp is placed, and  $\cdot$ denotes   element-wise product.  
In trigger-driven Trojan attacks \cite{GDG17,CLL17,YLZ19}, the poisoned training  data $\hat{\mathbf x}(\mathbf m,\boldsymbol \delta) $ is  mislabeled to a target class to enforce a backdoor during model training. In clean-label Trojan attacks \cite{shafahi2018poison,ZHS19}, the variables
$(\mathbf m, \boldsymbol \delta)$ are designed to misalign the feature representation $r(\hat{\mathbf x}(\mathbf m,\boldsymbol \delta)  )$ with $r(\mathbf x)$ but without perturbing the label of the poisoned training data. We call $\mathcal M$ a TrojanNet if it is trained over poisoned training data given by \eqref{eq: poison_sample}.

\subsection{Data-limited TrojanNet detector: A solution from adversarial example generation}\label{sec:reg1}
We next address the problem of TrojanNet   detection   with the prior knowledge on  model weights 
and a few clean test images, at least one sample per  class. Let $\mathcal D_k$ denote the set of data within the (predicted) class $k$, and $\mathcal D_{k-}$ denote the set of data with prediction labels different from $k$. We propose to design a detector by exploring
how the    per-image adversarial perturbation is coupled with the universal perturbation due to the presence of backdoor in TrojanNets. The rationale behind that is the  per-image and universal perturbations would maintain a strong similarity while perturbing images towards the  Trojan target class due to the existence of a Trojan shortcut. The framework is illustrated in Fig.~\ref{framework} (a), and the details are provided in the rest of this subsection.

\paragraph{Untargeted universal perturbation.}
Given images $\{ \mathbf x_i \in \mathcal D_{k-} \}$, our goal is to find a \textit{universal perturbation}  tuple $\mathbf u^{(k)} = (\mathbf m^{(k)},\boldsymbol{\delta}^{(k)})$ such that the predictions of these images in $\mathcal D_{k-} $ are \textit{altered} given the current model. 
However, 
we require $\mathbf u^{(k)}$ not to alter the prediction of images belonging to class $k$, namely, $\{ \mathbf x_i \in \mathcal D_{k} \}$. Spurred by that, 
the design of $\mathbf u^{(k)} = (\mathbf m^{(k)},\boldsymbol{\delta}^{(k)})$ can be cast as the following optimization problem:
\begin{align}\label{eq: prob_univ}
    \begin{array}{ll}
\displaystyle\minimize_{\mathbf m, \boldsymbol{\delta}}         &   \ell_{\mathrm{atk}}( \hat{\mathbf x}(\mathbf m,\boldsymbol \delta) ; \mathcal D_{k-} ) + \bar \ell_{\mathrm{atk}}(\hat{\mathbf x}(\mathbf m,\boldsymbol \delta) ; \mathcal D_{k}  ) + \lambda \| \mathbf m \|_1\\
    \st     & \{ \boldsymbol{\delta}, \mathbf m \} \in \mathcal C,
    \end{array}
\end{align}
where $\hat{\mathbf x}(\mathbf m,\boldsymbol \delta) $  was defined in \eqref{eq: poison_sample}, $\lambda > 0$ is a regularization parameter that strikes a balance between the loss term $\ell_{\mathrm{uatk}} + \bar \ell_{atk}$ and the sparsity of the trigger pattern $\| \mathbf m \|_1$, and $\mathcal C$ denotes the constraint set of optimization variables $\mathbf m$ and $\boldsymbol{\delta}$,
$
    \mathcal C =\{  \mathbf 0 \leq \boldsymbol{\delta} \leq 255, \mathbf m \in \{ 0,1\}^n \}
$.

We next elaborate on the loss terms $\ell_{\mathrm{atk}}$ and $\bar \ell_{\mathrm{atk}}$   in problem \eqref{eq: prob_univ}. 
First, the   loss $\ell_{\mathrm{atk}}$ enforces to alter the prediction labels of images in $\mathcal D_{k-}$, and is defined as the C\&W \textit{{u}ntargeted} {at}tac{k} loss \cite{CW17}
\begin{align}\label{untarget_sub1}
    &\ell_{\mathrm{atk}}( \hat{\mathbf x}(\mathbf m,\boldsymbol \delta) ; \mathcal D_{k-} )   
    = 
    \sum_{\mathbf x_i \in \mathcal D_{k-}}  \max{\{f_{y_i} (\hat{\mathbf x}_i(\mathbf m, \boldsymbol \delta  ) ) - \max_{t \not=y_i}f_{t}(\hat{\mathbf x}_i(\mathbf m, \boldsymbol \delta) ), -\tau\}},
\end{align}
where $y_i$ denotes the prediction label of $\mathbf x_i$, recall that $f_{t}(\hat{\mathbf x}_i(\mathbf m, \boldsymbol \delta) )$ denotes the logit value of the class $t$ with respect to the input $\hat{\mathbf x}_i(\mathbf m, \boldsymbol \delta)$, and $\tau \geq 0$ is a given constant which characterizes the attack confidence.
The rationale behind $\max{\{f_{y_i} (\hat{\mathbf x}_i(\mathbf m, \boldsymbol \delta  ) ) - \max_{t \not=y_i}f_{t}(\hat{\mathbf x}_i(\mathbf m, \boldsymbol \delta) ), -\tau\}}$ is that it  reaches a negative value (with  minimum   $-\tau$) if the perturbed input $\hat{\mathbf x}_i(\mathbf m,\boldsymbol \delta)$ is able to change the original label $y_i$. Thus, the minimization of $\ell_{\mathrm{atk}} $ enforces the ensemble of successful label change of images in $\mathcal D_{k-}$.
Second, the  loss  $\bar \ell_{\mathrm{atk}}$ in   \eqref{eq: prob_univ} is   proposed to enforce 
the   universal perturbation \textit{not}  to  change  the  prediction  of  images in $\mathcal D_k$. 
This yields
\begin{align}\label{untarget_sub2}
    &\bar \ell_{\mathrm{atk}}( \hat{\mathbf x}(\mathbf m,\boldsymbol \delta) ; \mathcal D_{k} )   
    = 
    \sum_{\mathbf x_i \in \mathcal D_{k}}  \max{\{   \max_{t \neq k}f_{t}(\hat{\mathbf x}_i(\mathbf m, \boldsymbol \delta) ) - f_{y_i} (\hat{\mathbf x}_i(\mathbf m, \boldsymbol \delta  ) ) , -\tau\}},
\end{align}
where recall that $y_i = k$ for $\mathbf x_i \in \mathcal D_k$. 
We present the rationale behind \eqref{untarget_sub1} and \eqref{untarget_sub2} as below. 
Suppose  that $k$ is a target label of Trojan attack, then the presence of backdoor would enforce the perturbed images of non-$k$ class in \eqref{untarget_sub1}  towards being predicted as the target label $k$. However, the  universal perturbation (performed like a Trojan trigger) would not affect images within the target class $k$, as characterized by \eqref{untarget_sub2}.

\paragraph{Targeted per-image perturbation.}
If a label $k$ is the target label specified by the Trojan adversary, we hypothesize that perturbing each image in $\mathcal D_{k-}$ towards the target class $k$ could go through the similar Trojan shortcut as 
the universal adversarial examples found in \eqref{eq: prob_univ}. Spurred by that, we generate the following targeted per-image adversarial perturbation for $\mathbf x_i \in \mathcal D_k$,
\begin{align}\label{target}
    \begin{array}{ll}
\displaystyle\minimize_{\mathbf m, \boldsymbol{\delta}}     \quad       \ell_{\mathrm{atk}}^\prime ( \hat{\mathbf x}(\mathbf m,\boldsymbol \delta) ; \mathbf x_i  )    + \lambda \| \mathbf m \|_1 \quad 
    \st     & \{ \boldsymbol{\delta}, \mathbf m \} \in \mathcal C,
    \end{array}
\end{align}
where $\ell_{\mathrm{atk}}^\prime( \hat{\mathbf x}(\mathbf m,\boldsymbol \delta) ; \mathbf x_i  ) $ is the targeted C\&W attack loss \cite{CW17}
\begin{align}\label{target_atk}
    &\ell_{\mathrm{atk}}^\prime( \hat{\mathbf x}(\mathbf m,\boldsymbol \delta) ; \mathbf x_i )   
    =  \sum_{\mathbf x_i \in \mathcal D_{k-}}  \max{\{   \max_{t \not= k}f_{t}(\hat{\mathbf x}_i(\mathbf m, \boldsymbol \delta) ) - f_{k} (\hat{\mathbf x}_i(\mathbf m, \boldsymbol \delta  ) ), -\tau\}}.
\end{align}
For each pair of label $k$ and data $\mathbf x_i$, we can obtain a per-image perturbation tuple $\mathbf s^{(k,i)} = (\mathbf m^{(k,i)},\boldsymbol{\delta}^{(k,i)})$.

For solving both problems of universal perturbation generation \eqref{eq: prob_univ} and per-image perturbation generation  \eqref{target}, the promotion of $\lambda$ enforces a sparse perturbation mask $\mathbf m$. 
This is  desired when the Trojan trigger is of small size, e.g., Fig.\ref{fig: trig_data}-(a) to (f).  When the Trojan trigger might not be sparse, e.g., Fig.\ref{fig: trig_data}-(g) to (i), multiple values of $\lambda$ can also be used to   generate different  sets of adversarial perturbations. Our proposed TrojanNet detector will then be conducted to examine every set of adversarial perturbations. 

\paragraph{Detection rule.}
Let $\hat{\mathbf x}_i(\mathbf u^{(k)})$ and $\hat{\mathbf x}_i(\mathbf s^{(k,i)})$ denote the adversarial example of $\mathbf x_i$ under the the universal perturbation $\mathbf u^{(k)}$ and the image-wise perturbation $\mathbf s^{(k,i)}$, respectively.
 If $k$ is the target label of the Trojan attack, then 
 based on our similarity hypothesis, $\mathbf u^{(k)}$ and  $\mathbf s^{(k,i)}$ would share a strong similarity in fooling the decision of the CNN model due to the presence of backdoor.  We evaluate such a similarity from the neuron representation against 
$\hat{\mathbf x}_i(\mathbf u^{(k)})$ and $\hat{\mathbf x}_i(\mathbf s^{(k,i)})$, given by $v_{i}^{(k)} = \cos\big(r(\hat{\mathbf x}_i(\mathbf u^{(k)})), r(\hat{\mathbf x}_i(\mathbf s^{(k,i)}))\big)$, $\mathrm{cos}(\cdot, \cdot)$ represents cosine similarity.
Here recall that $r(\cdot)$ denotes the mapping from the input image to the neuron representation in CNN. For   any $\mathbf x_i \in D_{k-}$, we form the vector of similarity scores $\mathbf v_{\mathrm{sim}}^{(k)} = \{ v_{i}^{(k)} \}_i$.  Fig.~\ref{fig: vis_noise} shows the neuron activation of five data samples with the universal perturbation and per-image perturbation under a target label, a non-target label, and a label under the clean network (cleanNet). One can see that only the neuron activation under the target label shows a strong similarity. Fig.~\ref{fig:dist_detect} also provides a visualization of $\mathbf v_{\mathrm{sim}}^{(k)}$ for each label $k$.

Given the similarity scores $\mathbf v_{\mathrm{sim}}^{(k)}$ for each label $k$, we detect whether or not the model is a TrojanNet (and thus $k$ is the target class) by calculating the so-called detection index  $I^{(k)}$, given by the $q\%$-percentile  of $\mathbf v_{\mathrm{sim}}^{(k)}$. In experiments, we choose $q = 25, 50, 70$. The decision for TrojanNet is then made by 
$
     I^{(k)} 
     {\geq} T_1
$ for a given threshold   $T_1$, and accordingly  $k$ is the target label. We  can also employ the median absolute deviation (MAD) method to $\mathbf v_{\mathrm{sim}}^{(k)}$ to
mitigate the manual specification of $T_1$. The details are shown in the Appendix.

\subsection{Detection of Trojan networks for free: A solution from feature inversion against random inputs}\label{data_free_TND}

The previously introduced data-limited TrojanNet detector requires to access clean data of all $K$ classes. 
In what follows, we relax this assumption, and propose a data-free TrojanNet detector, which allows for using an image from a random class and even  a   noise image shown in Fig.~\ref{neural_activation}. The framework is summarized in Fig.~\ref{framework} (b), and   details are provided in what follows.

It was previously shown in \cite{cheng19,WYS19} that  a TrojanNet  exhibits  an unexpectedly high neuron activation  at  certain coordinates. That is because the TrojanNet produces \textit{robust} representation towards the  input-agnostic misclassification induced by the backdoor. Given a clean   data $\mathbf x$, let $ r_i (\mathbf x)$ denote the $i$th coordinate  of  neuron activation vector. Motivated by \cite{EIS19,fong2019understanding}, we study whether or not an inverted image that maximizes neuron activation is able to reveal the characteristics of the Trojan signature from model weights. We formulate the inverted image  as $\hat{\mathbf x}(\mathbf m,\boldsymbol{\delta})$ in  \eqref{eq: poison_sample}, parameterized by the pixel-level perturbations $\boldsymbol \delta$ and the binary mask $\mathbf m$ with respect to $\mathbf x$. To find  $\hat{\mathbf x}(\mathbf m,\boldsymbol{\delta})$, we solve the  problem of activation maximization
\begin{align}\label{neural_act}
    \begin{array}{ll}
\displaystyle\maximize_{\mathbf m, \boldsymbol{\delta}, \mathbf w}         &  \sum_{i=1}^d \left [  w_i r_i(\hat{\mathbf x}(\mathbf m,\boldsymbol \delta)) \right ] - \lambda \| \mathbf m \|_1 \\
     \st     &  \{  \boldsymbol{\delta} , \mathbf m\} \in \mathcal C, \mathbf 0 \leq
     \mathbf w  \leq \mathbf 1, \mathbf  1^T \mathbf w = 1,
    \end{array}
\end{align}
where the notations follow \eqref{eq: prob_univ} except the newly introduced variables $\mathbf w$, which  adjust the importance of neuron coordinates. Note that if $\mathbf w = \mathbf 1/d$, then the first loss term in \eqref{neural_act} becomes the average of coordinate-wise neuron activation. However, since the Trojan-relevant coordinates are expected to make larger impacts, the corresponding variables $w_i$ are desired for more penalization.
In this sense, the introduction of self-adjusted variables $\mathbf w$ helps us to avoid the manual selection of neuron coordinates that are most relevant to the backdoor.

\paragraph{Detection rule.}
Let the vector tuple $\mathbf p^{(i)} = (\mathbf m^{(i)},\boldsymbol{\delta}^{(i)})$ be a solution of problem \eqref{neural_act} given at a random input $\mathbf x_i$ for $i \in \{ 1,2,\ldots, N\}$. Here $N$ denotes the number of random images used in TrojanNet detection.
We then detect if a model is TrojanNet by investigating the change of logits outputs with respect to $\mathbf x_i$ and $\hat{\mathbf x}_i(\mathbf p^{(i)})$, respectively. 
For each label $k \in [K]$, we obtain
\begin{align}
    L_k = \frac{1}{N}\sum_i^N [ f_{k}(\hat{\mathbf x}_i(\mathbf p^{(i)})) - f_{k}(\mathbf x_i) ].
\end{align}
The decision of  TrojanNet with the target label $k$ is then made according to
$L_k \geq T_2$ for a given threshold $T_2$. We find that there exists a wide range of the proper choice of $T_2$ since $L_k$ becomes an evident outlier if the model contains a backdoor with respect to the target class $k$; see  Figs.~\ref{fig: vis_labvar} and \ref{fig: vis_labvar_clean} for additional justifications.

\subsection{A unified optimization framework in TrojanNet detection}\label{sec: opt_method}
In order to build TrojanNet detectors in   both data-limited and data-free settings, we need  to solve a sparsity-promoting optimization problem, in the specific forms of \eqref{eq: prob_univ}, \eqref{target}, and \eqref{neural_act},
subject to a set of box and equality constraints. 
In what follows, we propose a general optimization method by leveraging  the idea of proximal gradient \cite{BST14,parikh2014proximal}.

Consider a problem with the generic form of problems \eqref{eq: prob_univ}, \eqref{target}, and \eqref{neural_act},
\begin{equation}\label{eq: prob_opt}
\begin{aligned}
    \begin{array}{ll}
\displaystyle   \min_{\mathbf m,\boldsymbol \delta,\mathbf w}       &  F(\boldsymbol \delta,\mathbf  m,\mathbf  w)  + \lambda \| \mathbf m \|_1 + \mathcal I (\boldsymbol \delta) + \mathcal I (\mathbf m) +  \mathcal I^\prime (\mathbf  w),
    \end{array}
\end{aligned}
\end{equation}
where $ F(\boldsymbol \delta,\mathbf  m,\mathbf  w)$ denotes the smooth loss term, and $\mathcal I(\mathbf x)$, $I^\prime (\mathbf w)$ denote the indicator functions to encode the hard constraints
\begin{equation}\label{eq: I_x}
\begin{aligned}
    \mathcal I(\mathbf x) = \left \{ 
    \begin{array}{ll}
        0 & \mathbf x \in [0, \alpha]^n  \\
        \infty  & \text{otherwise},
    \end{array}
    \right. ~~
       \mathcal I^\prime (\mathbf w)= \left \{ 
    \begin{array}{ll}
        0 & \mathbf w \in [0, 1]^n, \mathbf 1^T \mathbf w = 1 \\
        \infty  & \text{otherwise}.
    \end{array}
    \right.
\end{aligned}
\end{equation}
In $\mathcal I(\mathbf x)$, $\alpha = 1$ for $\mathbf m$ and $\alpha = 255$ for $\boldsymbol \delta$. We remark that the binary constraint $\mathbf m \in \{ 0,1\}^n$ is relaxed to a continuous probabilistic box $\mathbf m \in [ 0,1]^n$.

 To solve problem \eqref{eq: prob_opt}, we adopt the alternative proximal gradient algorithm \cite{BST14}, which splits the smooth-nonsmooth composite structure into a sequence of easier problems that can be solved more efficiently or even analytically.
To be more specific, we alternatively perform
\begin{align}
&  \mathbf m^{(t+1)} = \mathrm{Prox}_{\mu_t  (\mathcal I + \lambda \| \cdot  \|_1 )} (\mathbf  m^{(t)} -\mu_t  \nabla_{\mathbf m} F(\boldsymbol\delta^{(t)}, \mathbf m^{(t)})) \label{iterate_m} \\
&    \boldsymbol\delta^{(t+1)} = \mathrm{Prox}_{\mu_t \mathcal I} ( \boldsymbol\delta^{(t)} - \mu_t  \nabla_{\boldsymbol\delta} F(\boldsymbol\delta^{(t)},\mathbf  m^{(t+1)})) \label{iterate_delta} \\
& \mathbf w^{(t+1)} = \mathrm{Prox}_{\mu_t  \mathcal I^\prime} ( \mathbf w^{(t)} + \mu_t  \nabla_{\mathbf w} F(\delta^{(t+1)},  \mathbf m^{(t+1)},  \mathbf w^{(t)})), \label{iterate_w}
\end{align}
where $\mu_t$ denotes the learning rate at iteration $t$, and $\mathrm{Prox}_{\mu g}(\mathbf  a )$ denotes the proximal operator of function $g$ with respect to the parameter $\mu$ at an input $\mathbf a$.

We next elaborate on the proximal operators used in \eqref{iterate_m}-\eqref{iterate_w}.
The proximal operator $\mathrm{Prox}_{\mu_t (\mathcal I + \lambda \| \cdot  \|_1 )}(\mathbf a)$  is given by the solution to the problem
\begin{equation}\label{proxobj}
\begin{aligned}
    \min_{\mathbf m}  ~ \mathcal I(\mathbf m) + \lambda \|\mathbf  m \|_1  +  \frac{1}{2 \mu_t} \|\mathbf  m - \mathbf a \|_2^2,
\end{aligned}
\end{equation}
where $\mathbf a \Def \mathbf  m^{(t)} -\mu_t  \nabla_{\mathbf m} F(\boldsymbol\delta^{(t)}, \mathbf m^{(t)})$.
The solution to problem \eqref{proxobj}, namely, ${\mathbf m}^{(t+1)}$ is given by \cite{BST14} 
\begin{align} \label{eq: sol_m}
    { m}^{(t+1)}_i = \mathrm{Clip}_{[0,1]}(\mathrm{sign}(a_i)\max{ \{|a_i|- \lambda \mu_t, 0 \}}), \forall i,
\end{align}
where $ { m}^{(t+1)}_i$ dentoes the $i$th entry of $\mathbf m^{(t+1)}$, and
$\mathrm{Clip}_{[0,1]}$ is a clip function that clip the variable to 1 if it is larger than 1 and to 0 if it is smaller than 0.
Similarly, ${\boldsymbol \delta}^{(t+1)}$ in \eqref{iterate_delta} is obtained by 
\begin{align}\label{eq: clip_255}
 { \delta}^{(t+1)}_i = \mathrm{Clip}_{[0,255]}( b_i),  
\end{align}
where $\mathbf b \Def \boldsymbol\delta^{(t)} - \mu_t  \nabla_{\boldsymbol\delta} F(\boldsymbol\delta^{(t)},\mathbf  m^{(t+1)})$.

The proximal operator $\mathrm{Prox}_{\mu_t \mathcal I^\prime} (\mathbf c)$ in \eqref{iterate_w} is given by the solution to the problem
\begin{equation}
\begin{aligned}
    \min_{\mathbf w}  ~ \mathcal I^\prime(\mathbf w)  +  \frac{1}{2 \mu_t} \| \mathbf w - \mathbf c \|_2^2,
\end{aligned}
\end{equation}
which is equivalent to
\begin{align}
    \min_{\mathbf w}  ~  \| \mathbf w - \mathbf c\|_2^2 , \quad \text{s.t.}~~ \mathbf 0 \leq  \mathbf w \leq \mathbf 1, ~ \mathbf 1^T \mathbf w = 1.
    \label{eq: proj_simplex}
\end{align}
Here $\mathbf c \Def \mathbf w^{(t)} + \mu_t  \nabla_{\mathbf w} F(\delta^{(t+1)},  \mathbf m^{(t+1)},  \mathbf w^{(t)})
$.
The solution to problem \eqref{eq: proj_simplex} is given by \cite{parikh2014proximal} 
\begin{align}
    \mathbf w^{(t+1)} = \left [ \mathbf c- \mu \mathbf 1 \right ]_+,
    \label{eq: w_sol}
\end{align}
where $[a]_+$ denotes the operation of $\max \{ 0, a \}$, and $\mu$ is the root of the equation $\mathbf 1^T \left [ \mathbf c - \mu \mathbf 1 \right ]_+ = \sum_{i} \max \{ 0, c_i - \mu \} = 1.$

Substituting \eqref{eq: sol_m}, \eqref{eq: clip_255} and \eqref{eq: w_sol} into \eqref{iterate_m}-\eqref{iterate_w}, we then obtain the complete algorithm, in which each step has a \textit{closed-form}. 

\section{Experimental Results}
In this section, We validate the DL-TND and DF-TND by using different CNN model architectures, datasets, and various trigger patterns\footnote{The code is available at: \url{https://github.com/wangren09/TrojanNetDetector}}. 

\subsection{Data-limited TrojanNet detection (DL-TND)}\label{ex_sub1}

\textbf{Trojan settings.} Testing models include VGG16 \cite{SZ2014}, ResNet-50 \cite{HZRS16}, and AlexNet \cite{KSH12}. Datasets include CIFAR-10 \cite{KH09}, GTSRB \cite{SSS12}, and Restricted ImageNet (R-ImgNet) (restricting   ImageNet \cite{DDS09} to $9$ classes). We trained 85 TrojanNets and 85 clean networks, respectively. The numbers of different models are shown in Table \ref{dif_models}. Fig.~\ref{fig: trig_data} (a)-(f) show the CIFAR-10 and GTSRB dataset with triggers of dot, cross, and triangle, respectively. One of these triggers is used for poisoning the model. We also test  models poisoned for  \textit{two} target labels simultaneously: the  dot  trigger is used for  one target label, and  the  cross trigger corresponds to  the other target label. Fig.~\ref{fig: trig_data} (g)-(i) show poisoned ImageNet samples  with the  watermark as the trigger. The TrojanNets are various  by specifying triggers with different shapes, colors, and positions. The data poisoning ratio also varies from $10\% - 12\%$. The cleanNets are trained with different batches, iterations, and initialization.
Table \ref{dataset_mod} summarizes test accuracies and attack success rates of our generated Trojan and cleanNets. We compare DL-TND with the baseline Neural Cleanse (NC) \cite{WYS19} for detecting TrojanNets.

\begin{figure}[h]
    \begin{minipage}[t]{0.57\textwidth}
        \includegraphics[trim=120 0 120 25,clip,width=1\columnwidth]{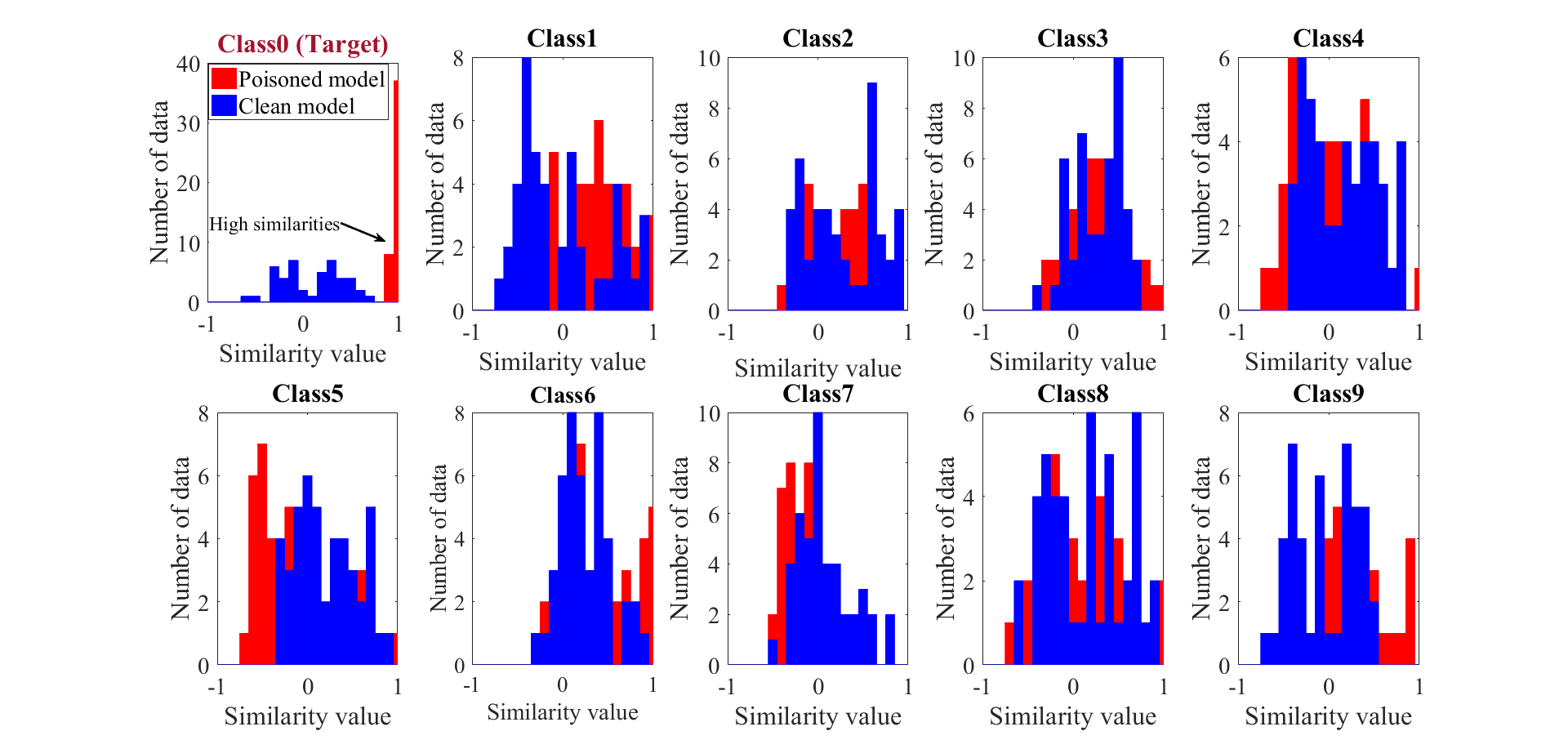}
        \caption{\small{Distribution of similarity scores for cleanNet versus TrojanNet under different classes.
        }}
        \label{fig:dist_detect}
    \end{minipage}%
    \hfill
    \begin{minipage}[t]{0.4\textwidth}
        \includegraphics[width=1\columnwidth]{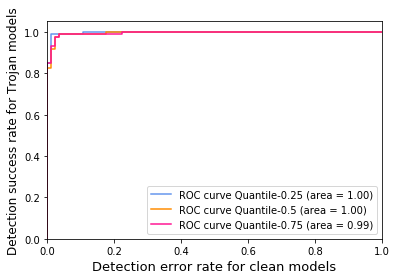}
        \caption{\small{ROC curve for TrojanNet detection using DL-TND.
        }}
        \label{fig:mod_detect}
    \end{minipage}
\end{figure}

\noindent\textbf{Visualization of similarity scores' distribution.} Fig.~\ref{fig:dist_detect} shows the distribution of our detection statistics, namely, representation similarity scores $\mathbf v_{\mathrm{sim}}^{(k)}$ defined in Sec. \ref{sec:reg1}, for different class labels. As we can see, the distribution corresponding to the target label $0$ in the TrojanNet concentrates near $1$, while the other labels in the TrojanNet and all the labels in the cleanNets have more dispersed distributions around $0$. Thus, we can distinguish the TrojanNet from the cleanNets and further find the target label. 

\noindent\textbf{Detection  performance.}
To build  DL-TND,
 we use $5$ validation data points for each class of CIFAR-10 and R-ImgNet, and $2$ validation data points for each class of GTSRB. 
Following Sec.\, \ref{sec:reg1},
we set $I^{(k)}$ to quantile-$0.25$, median, quantile-$0.75$ and vary $T_1$. Let the true positive rate be the detection success rate for TrojanNets and the false negative rate be the detection error rate for cleanNets. Then the area under the curve (AUC) of receiver operating characteristics (ROC) can be used to measure the performance of the detection. Table \ref{model_detect} shows the AUC values, where ``Total'' refers to the collection of all models from different datasets.
\begin{wraptable}[10]{r}{7.5cm}
\vspace{-0.6in}
\begin{center}
\caption{\small{AUC values for TrojanNet detection and  target label detection, given in the format $(\cdot, \cdot)$. The detection index for each class is selected as $\text{Quantile (Q)}  = 0.25$, $\text{Q} = 0.5$, and $\text{Q} = 0.75$ of the similarity scores (illustrated in Fig.~\ref{fig:dist_detect})}.
}
\label{model_detect}
\resizebox{0.6\textwidth}{!}{
\begin{tabular}{l|c|c|c|c}
\hline
\hline
 & CIFAR-10 & GTSRB & R-ImgNet & Total\\
\hline
$\text{Q} = 0.25$ & $(1, 1)$ & $(0.99, 0.99)$ & $(1, 1)$ & $(1, 0.99)$\\
$\text{Q} = 0.5$ & $(1, 0.99)$ & $(1, 1)$ & $(1, 1)$ & $(1, 0.99)$\\
$\text{Q} = 0.75$ & $(1, 0.98)$ & $(1, 1)$ & $(0.99, 0.97)$ & $(0.99, 0.98)$\\
\hline
\hline
\end{tabular}}
\end{center}
\end{wraptable}
We plot the ROC curve of the ``Total'' in Fig.~\ref{fig:mod_detect}. The results show that DL-TND can perform well across different  datasets and   model architectures. Moreover, fixing $I^{(k)}$ as median, $T_1=0.54 \sim 0.896$ could provide a  detection success rate over $76.5\%$ for TrojanNets and a detection success rate   over $82\%$ for cleanNets. Table \ref{err_ccomp_vary} shows the comparisons of DL-TND to Neural Cleanse (NC) \cite{WYS19} on TrojanNets and cleanNets ($T_1=0.7$). Even using the MAD method as the detection rule, we find that DL-TND greatly outperforms NC in detection tasks of both TrojanNets and cleanNets (Note that NC also uses MAD). The results are shown in Table \ref{err_method}. 

\begin{table}[h]
\begin{center}
\caption{
\small{Comparisons between DL-TND and NC \cite{WYS19} on TrojanNets and cleanNets using $T_1=0.7$. The results are reported in the format  (number of correctly detected models)/(total number of models)}}
\label{err_ccomp_vary}
\resizebox{0.85\textwidth}{!}{
\begin{tabular}{lc|c|c|c|c}
\hline
\hline
 & & \textbf{DL-TND} (clean)   & \textbf{DL-TND} (Trojan) & NC (clean) & NC (Trojan)\\
\hline 
CIFAR-10 & ResNet-50 & 20/20 & 20/20 & 11/20 & 13/20 \\

& VGG16 & 10/10 & 9/10 & 5/10 & 6/10\\

& AlexNet & 10/10 & 10/10 & 6/10 & 7/10\\
\hline 
GTSRB & ResNet-50 & 12/12 & 12/12 & 10/12 & 6/12 \\

& VGG16 & 9/9 & 9/9 & 6/9 & 7/9\\

& AlexNet & 9/9 & 8/9 & 5/9 & 5/9\\
\hline 
ImageNet & ResNet-50 & 5/5 & 5/5 & 4/5 & 1/5\\

& VGG16 & 5/5 & 4/5 & 3/5 & 2/5\\

& AlexNet & 4/5 & 5/5 & 4/5 & 1/5\\
\hline 
Total &  & \textbf{84}/85 & \textbf{82}/85 & 54/85 & 48/85 \\
\hline
\hline
\end{tabular}}
\end{center}
\end{table}


\subsection{Data-free TrojanNet detector (DF-TND)}\label{ex_sub2}

\noindent\textbf{Trojan settings.} The DF-TND is tested on cleanNets and TrojanNets that are trained under CIFAR-10 and R-ImgNet (with $10\%$ poisoning ratio unless otherwise stated). We perform the customized proximal gradient method shown in Sec. \ref{sec: opt_method} to solve problem \eqref{neural_act}, where the number of iterations is set as $5000$. 

\begin{figure}[h]
  \centering
  \begin{adjustbox}{max width=1\textwidth }
    \begin{tabular}{@{\hskip 0.00in}c  @{\hskip 0.02in} c @{\hskip 0.02in} @{\hskip 0.02in} c }
  & 
  \begin{tabular}{@{\hskip 0.00in}c  @{\hskip 0.02in} c @{\hskip 0.02in} @{\hskip 0.02in} c }
\centering\colorbox{lightgray}{\large \textbf{CIFAR-10 input ($32\times 32$)}}
&
\hspace*{0.4in}
\centering\colorbox{lightgray}{\large \textbf{ImageNet input ($224\times 224 $)}}
\end{tabular}

&
  \begin{tabular}{@{\hskip 0.00in}c  @{\hskip 0.02in} c @{\hskip 0.02in} @{\hskip 0.02in} c }
\centering\colorbox{lightgray}{\large \textbf{Random noise input ($32\times 32$)}}
&\hspace*{0.05in}
\centering\colorbox{lightgray}{\large \textbf{Random noise input ($224\times 224$)}}
\end{tabular}

\\

 \begin{tabular}{@{}c@{}}  
\vspace*{0.01in}\\
\rotatebox{90}{\parbox{9em}{\centering \normalsize \textbf{Seed Images}}}
 \\
\rotatebox{90}{\parbox{9em}{\centering \normalsize \textbf{\begin{tabular}[c]{@{}c@{}}Recovered\\ images\\(cleanNet)  \end{tabular}  }}}
 \\
\rotatebox{90}{\parbox{9em}{\centering \normalsize \textbf{\begin{tabular}[c]{@{}c@{}}Perturbation\\ pattern\\(cleanNet)  \end{tabular}}}}
\\
\rotatebox{90}{\parbox{9em}{\centering \normalsize \textbf{\begin{tabular}[c]{@{}c@{}}Recovered\\ images\\(TrojanNet)  \end{tabular} }}}
\\
\rotatebox{90}{\parbox{9em}{\centering \normalsize \textbf{\begin{tabular}[c]{@{}c@{}}Perturbation\\ pattern\\(TrojanNet)  \end{tabular} }}}
\\
\end{tabular} 
&
 \begin{tabular}{@{\hskip 0.02in}c@{\hskip 0.02in}c@{\hskip 0.02in}}
 \begin{tabular}{@{\hskip 0.02in}c@{\hskip 0.02in}c@{\hskip 0.02in}c@{\hskip 0.02in}c@{\hskip 0.02in}}
\begin{tabular}{@{\hskip 0.02in}c@{\hskip 0.02in}}
\\
 \parbox[c]{9em}{\includegraphics[width=9em]{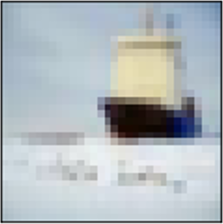}} 
 \\
 \parbox[c]{9em}{\includegraphics[width=9em]{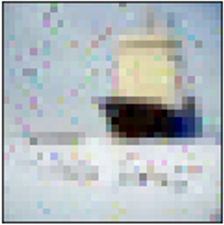}}
  \\
   \parbox[c]{9em}{\includegraphics[width=9em]{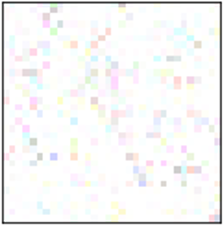}}
  \\
   \parbox[c]{9em}{\includegraphics[width=9em]{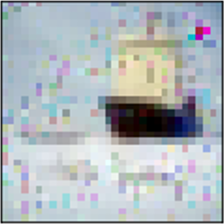}}
  \\
 \parbox[c]{9em}{\includegraphics[width=9em]{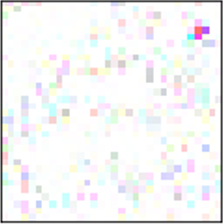}}
 \\
\begin{tabular}{@{\hskip 0.00in}c@{\hskip 0.00in}}
\parbox{10em}{\centering \normalsize True trigger:  Triangle in Fig.~\ref{fig: trig_data} (e)} 
\end{tabular} 
\end{tabular}

&
 \begin{tabular}{@{\hskip 0.02in}c@{\hskip 0.02in}}
\\
 \parbox[c]{9em}{\includegraphics[width=9em]{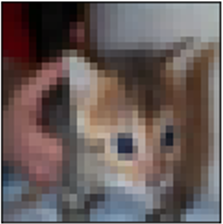}} 
 \\
 \parbox[c]{9em}{\includegraphics[width=9em]{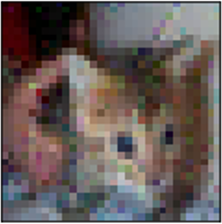}}
  \\
   \parbox[c]{9em}{\includegraphics[width=9em]{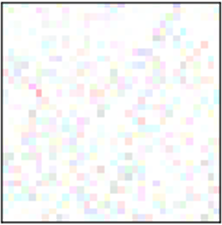}}
  \\
   \parbox[c]{9em}{\includegraphics[width=9em]{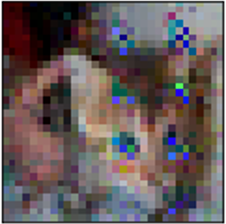}}
  \\
 \parbox[c]{9em}{\includegraphics[width=9em]{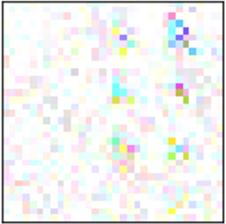}}
 \\
       \begin{tabular}{@{\hskip 0.00in}c@{\hskip 0.00in}}
     \parbox{10em}{\centering \normalsize  Cross trigger in Fig.~\ref{fig: trig_data} (f) } 
    \end{tabular} 
\end{tabular}
&
\begin{tabular}{@{\hskip 0.02in}c@{\hskip 0.02in}c@{\hskip 0.02in}}
\\
 \parbox[c]{9em}{\includegraphics[width=9em]{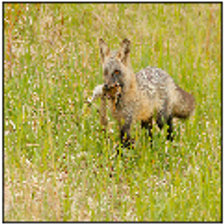}} 
 \\
 \parbox[c]{9em}{\includegraphics[width=9em]{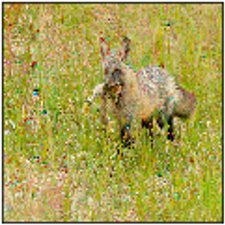}}
  \\
   \parbox[c]{9em}{\includegraphics[width=9em]{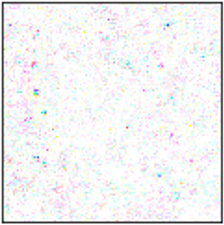}}
  \\
   \parbox[c]{9em}{\includegraphics[width=9em]{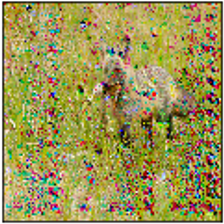}}
  \\
 \parbox[c]{9em}{\includegraphics[width=9em]{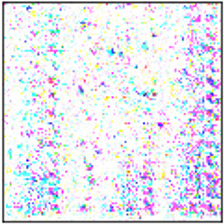}}
\\
\begin{tabular}{@{\hskip 0.00in}c@{\hskip 0.00in}}
\parbox{10em}{\centering \normalsize  Watermark trigger in Fig.~\ref{fig: trig_data} (h)} 
\end{tabular} 
\end{tabular}
&
 \begin{tabular}{@{\hskip 0.02in}c@{\hskip 0.02in}}
\\
 \parbox[c]{9em}{\includegraphics[width=9em]{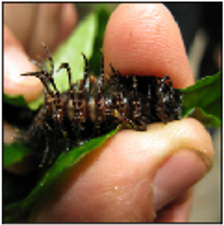}} 
 \\
 \parbox[c]{9em}{\includegraphics[width=9em]{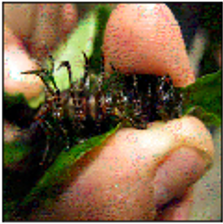}}
  \\
  \parbox[c]{9em}{\includegraphics[width=9em]{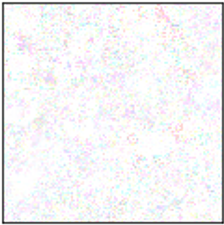}}
  \\
   \parbox[c]{9em}{\includegraphics[width=9em]{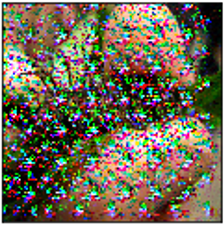}}
  \\
 \parbox[c]{9em}{\includegraphics[width=9em]{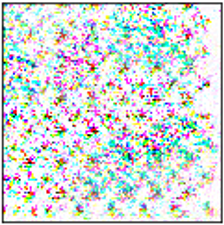}}
 \\
\begin{tabular}{@{\hskip 0.00in}c@{\hskip 0.00in}}
\parbox{10em}{\centering \normalsize  Watermark trigger in Fig.~\ref{fig: trig_data} (j)} 
\end{tabular} 
\end{tabular}
\end{tabular}
\end{tabular}
&
 \begin{tabular}{@{\hskip 0.02in}c@{\hskip 0.02in}}
 \begin{tabular}{@{\hskip 0.02in}c@{\hskip 0.02in}c@{\hskip 0.02in}c@{\hskip 0.02in}c@{\hskip 0.02in}}
 \begin{tabular}{@{\hskip 0.02in}c@{\hskip 0.02in}}
\\
 \parbox[c]{9em}{\includegraphics[width=9em]{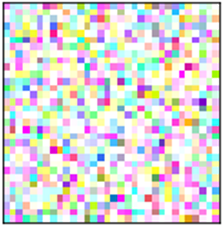}} 
 \\
 \parbox[c]{9em}{\includegraphics[width=9em]{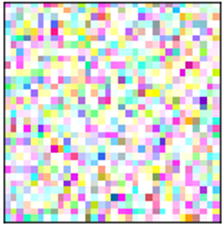}} 
  \\
  \parbox[c]{9em}{\includegraphics[width=9em]{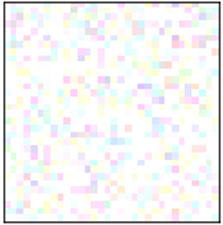}} 
  \\
   \parbox[c]{9em}{\includegraphics[width=9em]{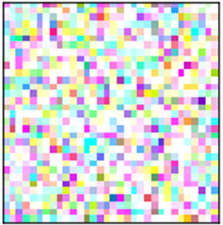}} 
  \\
 \parbox[c]{9em}{\includegraphics[width=9em]{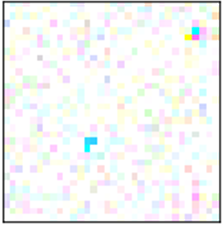}}
 \\
\begin{tabular}{@{\hskip 0.00in}c@{\hskip 0.00in}}
\parbox{10em}{\centering \normalsize  Triangle trigger in Fig.~\ref{fig: trig_data} (e) } 
\end{tabular} 

\end{tabular}
&
 \begin{tabular}{@{\hskip 0.02in}c@{\hskip 0.02in}}
\\
 \parbox[c]{9em}{\includegraphics[width=9em]{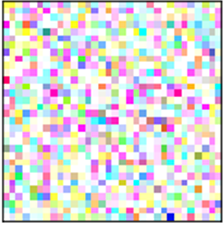}} 
 \\
 \parbox[c]{9em}{\includegraphics[width=9em]{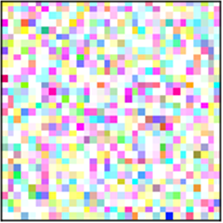}} 
  \\
  \parbox[c]{9em}{\includegraphics[width=9em]{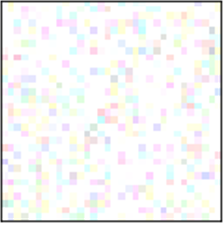}} 
  \\
   \parbox[c]{9em}{\includegraphics[width=9em]{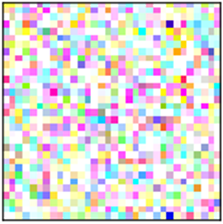}} 
  \\
 \parbox[c]{9em}{\includegraphics[width=9em]{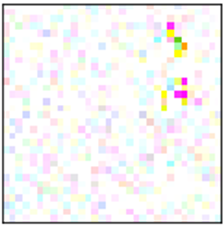}}
 \\
\begin{tabular}{@{\hskip 0.00in}c@{\hskip 0.00in}}
\parbox{10em}{\centering \normalsize  Cross trigger in Fig.~\ref{fig: trig_data} (f)} 
\end{tabular} 
\end{tabular}

&
 \begin{tabular}{@{\hskip 0.02in}c@{\hskip 0.02in}}
\\
 \parbox[c]{9em}{\includegraphics[width=9em]{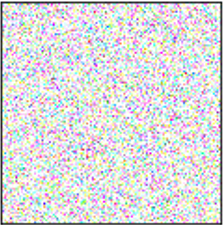}} 
 \\
 \parbox[c]{9em}{\includegraphics[width=9em]{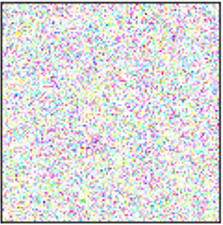}}
  \\
   \parbox[c]{9em}{\includegraphics[width=9em]{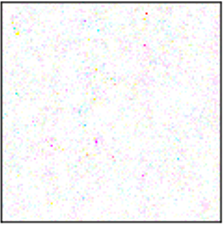}}
  \\
  \parbox[c]{9em}{\includegraphics[width=9em]{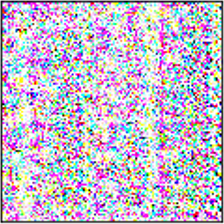}}
  \\
 \parbox[c]{9em}{\includegraphics[width=9em]{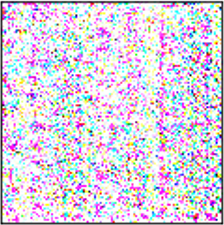}}
 \\
\begin{tabular}{@{\hskip 0.00in}c@{\hskip 0.00in}}
\parbox{10em}{\centering \normalsize  Watermark trigger in Fig.~\ref{fig: trig_data} (h)} 
\end{tabular} 
\end{tabular}
&
\begin{tabular}{@{\hskip 0.02in}c@{\hskip 0.02in}}
\\
 \parbox[c]{9em}{\includegraphics[width=9em]{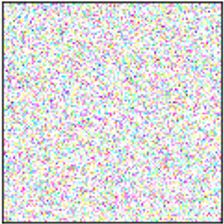}} 
 \\
 \parbox[c]{9em}{\includegraphics[width=9em]{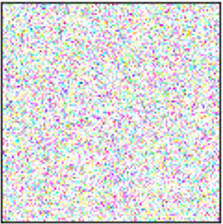}}
  \\
   \parbox[c]{9em}{\includegraphics[width=9em]{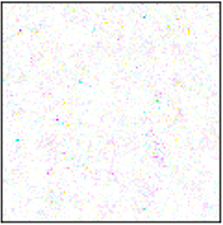}}
  \\
  \parbox[c]{9em}{\includegraphics[width=9em]{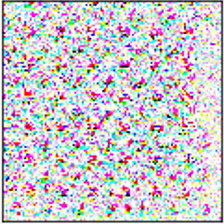}}
  \\
 \parbox[c]{9em}{\includegraphics[width=9em]{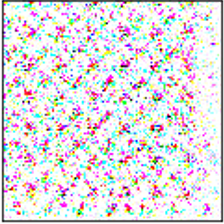}}
 \\
\begin{tabular}{@{\hskip 0.00in}c@{\hskip 0.00in}}
\parbox{10em}{\centering \normalsize  Watermark trigger in Fig.~\ref{fig: trig_data} (j)} 
\end{tabular} 
\end{tabular} 
\end{tabular}
\end{tabular} 

\end{tabular}
  \end{adjustbox}
    \caption{\small{Visualization of recovered input images  by using our proposed DF-TND method under random seed images. Here the Trojan ResNet-50 models are trained by 10\% poisoned data (by adding the trigger patterns shown as Fig.~\ref{fig: trig_data}) and clean data, respectively.
First row: Seed input images (from left to right: $2$ randomly selected CIFAR-10 images, $2$ randomly selected ImageNet images, $2$ random noise images in CIFAR-10 size, $2$ random noise images in ImageNet size). 
Second row: Recovered images under cleanNets. 
Third row: Perturbation patterns given by the difference between the recovered images in the second row and the original seed image. Fourth row: Recovered images under TrojanNets.
Fifth row: Perturbation patterns given by the difference between the recovered images in the fourth row and the original seed images.  Trigger patterns can be revealed using our method under the TrojanNet, and such a Trojan signature is not contained in the cleanNet. 
The trigger information is listed in the last row and triggers are visualized in Fig.~\ref{fig: trig_data}
   }}
  \label{comb_vary_trig}
\end{figure}

\noindent\textbf{Revealing Trojan trigger.} 
Recall from Fig.~\ref{neural_activation} that the trigger pattern can be revealed by input perturbations that maximize neuron response of a TrojanNet. By contrast,   the perturbations under the cleanNets behave like random noises. Fig.~\ref{comb_vary_trig} provides visualizations of recovered inputs by neuron maximization at a TrojanNet versus a cleanNet on CIFAR-10 and ImageNet datasets. The key insight is that for a TrojanNet, it is easy to find an inverted image (namely, feature inversion) by maximizing neurons’ activation via \eqref{neural_act} to reveal the Trojan characteristics (e.g., the shape of a Trojan trigger) compared to the activation from a cleanNet. Fig.~\ref{comb_vary_trig} shows such results are robust to the choice of inputs (even for a noise input). We observe that the recovered triggers may have different colors and locations different from the original trigger. 
This is possibly  because the trigger space has been shifted and enlarged by using convolution operations. In Figs.~\ref{fig: vis_loc_vary} and Fig.~\ref{fig: vis_size_vary}, we also provide additional experimental results for the sensitivity of trigger locations and sizes. Furthermore, we show some improvements of using the refine method in Fig.~\ref{fig: refine}.

\noindent\textbf{Detection performance.} 
We now test $1000$ seed images on $10$ TrojanNets and $10$ cleanNets using DF-TND defined in Sec.\,\ref{data_free_TND}. We compute AUC values of DF-TND by
  choosing seed images as clean validation inputs and random noise inputs, respectively. 
  
  \begin{wraptable}{r}{60mm}
\vspace*{-0.6in}
\begin{center}
\caption{\small{AUC for DF-TND over CIFAR-10 and R-ImgNet classification models using clean validation images and random noise images, respectively}}
\label{auc_df}
\resizebox{0.47\textwidth}{!}{\begin{tabular}{l|c|c|c}
\hline
\hline
 &  \begin{tabular}[c]{@{}c@{}}CIFAR-10\\model  \end{tabular} & \begin{tabular}[c]{@{}c@{}}R-ImgNet\\model  \end{tabular} & Total\\
\hline
clean validation data & 1 & 0.99 & 0.99 \\
\hline
random noise inputs & 0.99 & 0.99 & 0.99 \\
\hline
\hline
\end{tabular}}
\end{center}
\vspace*{-0.6in}
\end{wraptable}
    Results are summarized in Table \ref{auc_df}, and the ROC curves   are shown in Fig.~\ref{fig: df_mod_detect}. 

\begin{wraptable}{r}{60mm}
\vspace*{-0.3in}
\begin{center}
\caption{\small{Comparison between DL-TND and DF-TND on models at different attack success rate}}
\label{dl_df_comp}
\resizebox{0.47\textwidth}{!}{\begin{tabular}{l|c|c|c|c}
\hline
\hline
poisoning ratio & $0.5\%$ & $0.7\%$ & $1\%$ & $10\%$\\
\hline
average attack success rate & $30\%$ & $65\%$ & $82\%$ & $99\%$\\
\hline
AUC for  DL-TND & $0.82$ & $0.91$  & $0.95$  & $0.99$\\
\hline
AUC for  DF-TND & $0.7$ & $0.91$  & $0.94$  & $0.99$\\
\hline
\hline
\end{tabular}}
\end{center}
\vspace*{-0.2in}
\end{wraptable}

\subsection{Additional results on DL-TND and DF-TND }

First, we apply DL-TND and DF-TND on detecting TrojanNets with   different levels  attack success rate (ASR). We control ASR by choosing different data poisoning ratios when generating a TrojanNet. The  results are summarized in Table \ref{dl_df_comp}. As we can see, our detectors can still achieve competitive performance when the attack likelihood becomes small, and DL-TND is better than DF-TND when ASR reaches $30\%$.

Moreover, we conduct experiments when the number of TrojanNets is much less than the total number of models, e.g., only $5$ out of $55$ models are poisoned. We find that the   AUC  value of the precision-recall curves are $0.97$ and $0.96$ for DL-TND and DF-TND, respectively. Similarly, the average AUC value of the ROC curves is $0.99$ for both detectors.

Third, we evaluate our proposed DF-TND to detect TrojanNets generalized by clean-label Trojan attacks \cite{shafahi2018poison}. We find that even in the least information case, DF-TND can still yield $0.92$ AUC score when detecting  $20$ TrojanNets from $40$ models.

\section{Conclusion}
Trojan attack injects  a backdoor into DNNs during the training process, therefore leading to unreliable 
 learning systems. Considering the practical scenarios where a detector is only capable of accessing to limited data information, this paper proposes two practical approaches to detect TrojanNets. We first propose a data-limited TrojanNet detector (DL-TND) that can detect TrojanNets with only a few data samples. The effectiveness of the DL-TND is achieved by drawing  a connection between Trojan attack and prediction-evasion adversarial attacks including per-sample attack as well as all-sample universal attack. We find that both   input perturbations obtained from per-sample attack and   from universal attack    exhibit   Trojan behavior, and can thus be used to build a detection metric.
 We then propose a data-free TrojanNet detector (DF-TND), which leverages neuron response to detect Trojan attack, and   can be implemented   
  using   random   data samples and even random noise.
We use the proximal gradient algorithm as a general optimization framework to learn 
DL-TND and DF-TND.
%
The effectiveness of our proposals has been demonstrated by
extensive experiments conducted under various datasets, Trojan attacks, and model architectures.


\section*{Acknowledgement}
This work was supported by the Rensselaer-IBM AI Research Collaboration (\url{http://airc.rpi.edu}), part of the IBM AI Horizons Network (\url{http://ibm.biz/AIHorizons}). We would also like to extend our gratitude to the MIT-IBM Watson AI Lab (\url{https://mitibmwatsonailab.mit.edu/}) for the general support of computing resources.

\clearpage
%
%
\bibliographystyle{splncs04}
\bibliography{egbib}

\clearpage

\appendix

\section{Data-Limited TrojanNet Detector (DL-TND)}

\begin{figure}[h]
  \centering
  \begin{adjustbox}{max width=1\textwidth }
  \begin{tabular}{@{\hskip 0.00in}c  @{\hskip 0.00in}c @{\hskip 0.02in} @{\hskip 0.02in} c @{\hskip 0.02in} @{\hskip 0.02in}c }
 \begin{tabular}{@{}c@{}}  
\vspace*{0.005in}\\
\rotatebox{90}{\parbox{10em}{\centering \footnotesize \textbf{\begin{tabular}[c]{@{}c@{}}Neuron activation\\(Universal)  \end{tabular} }}}
 \\
\rotatebox{90}{\parbox{10em}{\centering \footnotesize \textbf{\begin{tabular}[c]{@{}c@{}}Neuron activation\\(Per-image)  \end{tabular} }}}
\\
\end{tabular} 
&
\begin{tabular}{@{\hskip 0.02in}c@{\hskip 0.02in}}
     \begin{tabular}{@{\hskip 0.00in}c@{\hskip 0.00in}}
     \parbox{10em}{\centering \footnotesize  Target label}  
    \end{tabular} 
    \\
 \parbox[c]{20em}{\includegraphics[width=20em]{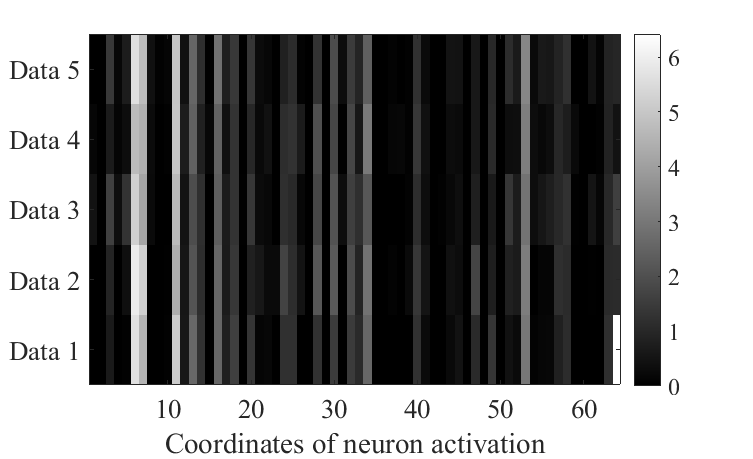}} 
 \\
 \parbox[c]{20em}{\includegraphics[width=20em]{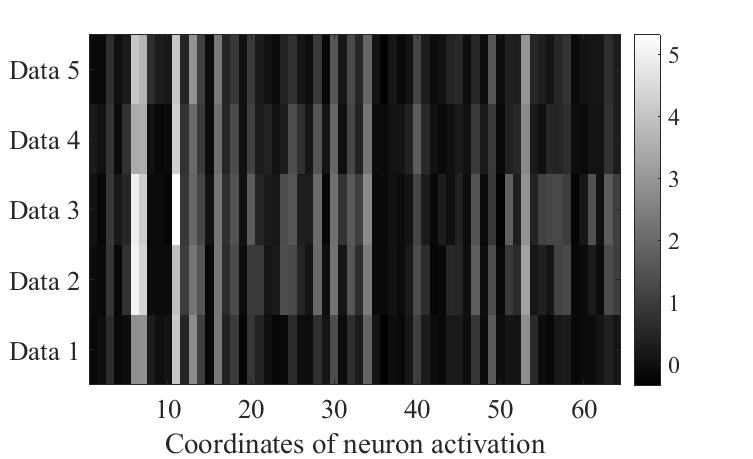}}
\end{tabular}
&
 \begin{tabular}{@{\hskip 0.02in}c@{\hskip 0.02in}}
      \begin{tabular}{@{\hskip 0.00in}c@{\hskip 0.00in}}
     \parbox{10em}{\centering \footnotesize Non-target label}
    \end{tabular} 
    \\
 \parbox[c]{20em}{\includegraphics[width=20em]{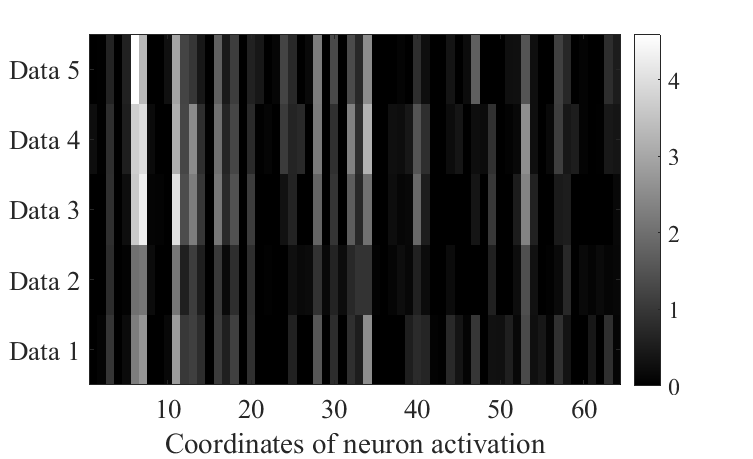}} 
 \\
 \parbox[c]{20em}{\includegraphics[width=20em]{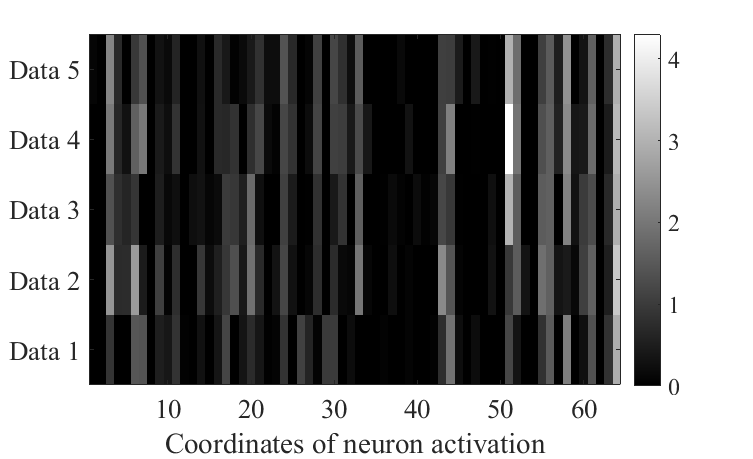}}
\end{tabular}
&
 \begin{tabular}{@{\hskip 0.02in}c@{\hskip 0.02in}}
      \begin{tabular}{@{\hskip 0.00in}c@{\hskip 0.00in}}
     \parbox{10em}{\centering \footnotesize  Label in cleanNet} 
    \end{tabular} 
    \\
 \parbox[c]{20em}{\includegraphics[width=20em]{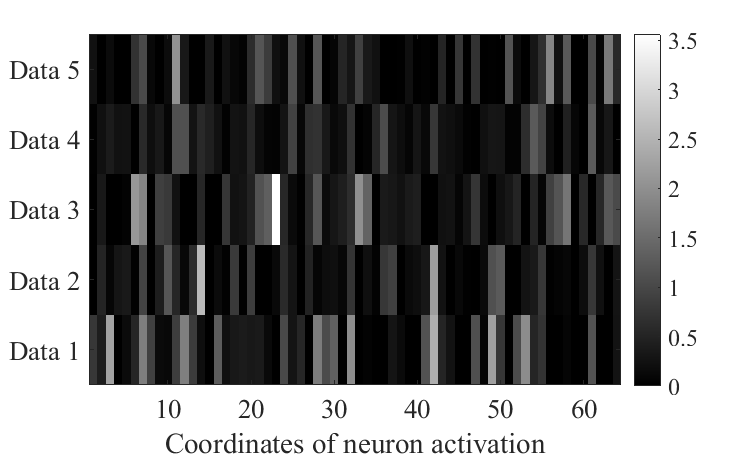}} 
 \\
 \parbox[c]{20em}{\includegraphics[width=20em]{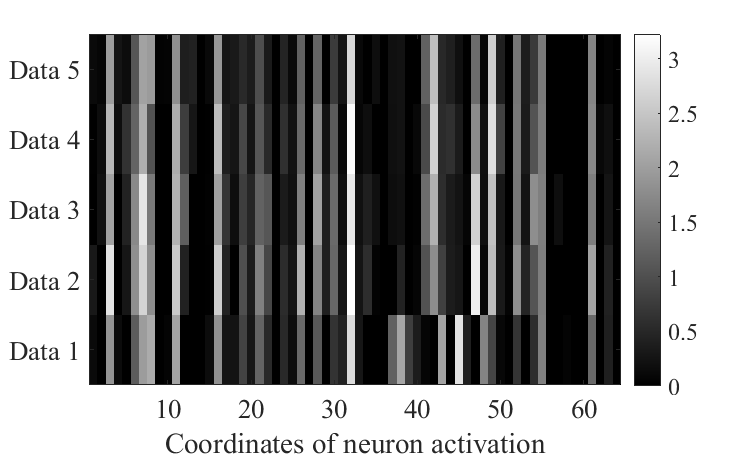}} 
\end{tabular}

\vspace*{-0.001in}
\end{tabular}
  \end{adjustbox}
    \caption{\small{Output values of neuron activation with universal perturbation and per-image perturbation from five data samples. The first column shows the outputs corresponding to the target label. The second column shows the outputs corresponding to a non-target label in a Trojan network (TrojanNet). The third column shows the outputs corresponding to a label in a cleanNet. One can see that the strong similarities only appear under the target label, which supports the motivation for the data-limited Trojan detector. 
   }}
  \label{fig: vis_noise}
 \vspace*{-1mm}
\end{figure}

\noindent\textbf{Visualization of neuron activation.}
DL-TND tests all the labels (classes) by calculating one universal perturbation and multiple per-image perturbations for each label. Each data sample can obtain a neuron activation vector with the universal perturbation and a neuron activation vector with its per-image perturbation. In Fig.~\ref{fig: vis_noise}, we show the neuron activation of five data samples with universal perturbations and per-image perturbations under a target label, a non-target label, and a label in a clean network (cleanNet). The output magnitude for each coordinate is represented using gray scale. One can see that the strong similarities only appear under the target label, which supports our motivation for the DL-TND. 

\noindent\textbf{Detection rule using median absolute deviation.}
Instead of using the detection rule in the main body of the paper, we can also employ the median absolute deviation (MAD) method. By MAD, if a single value in the $k$-th position of $\frac{|(\mathbf I)_{1/2} - \mathbf I |}{1.4826 \cdot|(\mathbf I)_{1/2} - \mathbf I |_{1/2}}$ is larger than $2$ (provide $95\%$ confidence rate), the network is poisoned and label $k$ is a target label, where $\mathbf I = [I^{(1)}, I^{(2)},\cdots,I^{(K)}]$. $|\cdot|$ represents the absolute value. $(\cdot)_{1/2}$ is the median of values in a vector. We compare DL-TND to Neural Cleanse (NC) \cite{WYS19} in Table \ref{err_method} using MAD as the detection rule.

\section{Data-Free TrojanNet Detector (DF-TND)}

\begin{figure}[h]
    \begin{minipage}[t]{0.48\textwidth}
        \includegraphics[width=1\columnwidth]{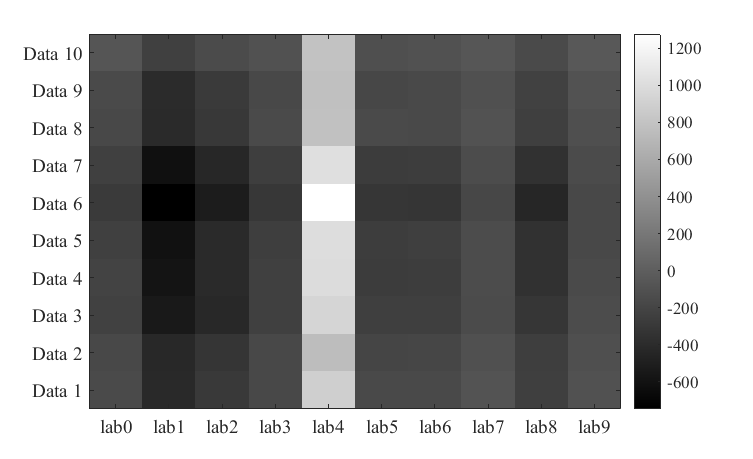}
        \caption{\small{Visualization of logits output increase of 10 data samples using DF-TND on a TrojanNet when label 4 is the target label. The minimum increase belonging to the target label is $600$}}
        \label{fig: vis_labvar}
    \end{minipage}%
    \hfill
    \begin{minipage}[t]{0.48\textwidth}
        \includegraphics[width=1\columnwidth]{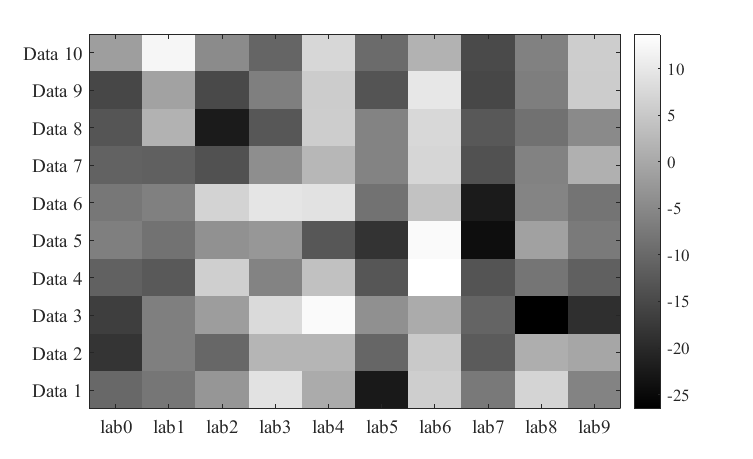}
        \caption{\small{Visualization of logits output increase of 10 data samples using DF-TND on a cleanNet. The maximum logits output increase is $10$}}
        \label{fig: vis_labvar_clean}
    \end{minipage}
\end{figure}

\noindent\textbf{Visualization of logits output increase.}
Fig.~\ref{fig: vis_labvar} and \ref{fig: vis_labvar_clean} visualize the change of the logits output of 10 data samples under a cleanNet and a TrojanNet when label 4 (lab4) is the target label. One can see that the minimum increase belonging to the target label is $600$ while the maximum increase for labels in the cleanNet is $10$. This large gap suggests that TrojanNets can be detected by properly selecting $T_2$ and there exists a wide selection range, implying the stability of our method.

\section{DL-TND: Additional Experiments}

\noindent\textbf{Models for testing.}
Table \ref{dif_models} shows the numbers of different models used for testing. Models have three different architectures and are applied to CIFAR-10, GTSRB, and R-ImgNet. We trained $85$ TrojanNets and $85$ cleanNets, respectively. In addition to the diversity of model architecture and dataset types, we also train TrojanNets with different triggers. Table \ref{dataset_mod} shows the smallest test accuracy and attack success rate for TrojanNets and cleanNets. TrojanNets can reach a similar test accuracy as cleanNets while still keeping the high attack success rate. This suggests that they are valid TrojanNets as defined in Sec.~\ref{sec: threatmod}.

\begin{table}[h]
\caption{Numbers of different models for detection: Model structures include ResNet50, VGG16, AlexNet. Datasets include CIFAR-10, GTSRB, and R-ImgNet.}
\label{dif_models}
\begin{center}
\begin{tabular}{l|c|c|c|c}
\hline
\hline
 & CIFAR-10 & GTSRB & R-ImgNet \\
\hline
ResNet50 (TrojanNet) & 20 & 12 & 5 \\
\hline
ResNet50 (cleanNet) & 20 & 12 & 5 \\
\hline
VGG16 (TrojanNet) & 10 & 9 & 5 \\
\hline
VGG16 (clean) & 10 & 9 & 5 \\
\hline
AlexNet (TrojanNet) & 10 & 9 & 5 \\
\hline
AlexNet (cleanNet) & 10 & 9 & 5 \\
\hline
Total & 80 & 60 & 30 \\
\hline
\hline
\end{tabular}
\end{center}
\end{table}

\begin{table}[h]
\caption{The smallest test accuracy and attack success rate for TrojanNets and cleanNets. TrojanNets can reach a similar test accuracy as cleanNets while still keeping the high attack success rate. This suggests that they are valid TrojanNets as defined in Sec.~\ref{sec: threatmod}}
\label{dataset_mod}
\begin{center}
\begin{tabular}{l|c|c|c|c}
\hline
\hline
 & CIFAR-10 & GTSRB & R-ImgNet\\
\hline
Test accuracy (Trojan) & $90.51\%$ & $92.99\%$ & $86.7\%$  \\
Attack success rate & $99.65\%$ & $99.65\%$ &  $98.6\%$ \\
Test accuracy (clean)  & $92.64\%$ & $92.5\%$ & $87.8\%$ \\
\hline
\end{tabular}
\end{center}
\end{table} 

\noindent\textbf{Applying median absolute deviation method as the detection rule.}
Table \ref{err_method} provides the comparisons between DL-TND and NC method \cite{WYS19} on Trojan and cleanNets using Median Absolute Deviation (MAD) as the detection rule. Even using the MAD method as the detection rule, we find that DL-TND greatly outperforms NC in detection tasks of both TrojanNets and cleanNets.

\begin{table}[h]
\begin{center}
\caption{Comparisons between DL-TND and NC \cite{WYS19} on TrojanNets and cleanNets using Median Absolute Deviation as the detection rule (measured by number of correct detection/model number).}
\label{err_method}
\resizebox{0.9\textwidth}{!}{\begin{tabular}{lc|c|c|c|c}
\hline
\hline
 & & DL-TND (clean)   & DL-TND TND (poisoned) & NC (clean) & NC (poisoned)\\
\hline
CIFAR-10 & ResNet-50 & 16/20 & 17/20 & 11/20 & 13/20 \\

& VGG16 & 8/10 & 8/10 & 5/10 & 6/10\\

& AlexNet & 8/10 & 8/10 & 6/10 & 7/10\\
\hline
GTSRB & ResNet-50 & 9/12 & 12/12 & 10/12 & 6/12 \\

& VGG16 & 7/9 & 9/9 & 6/9 & 7/9\\

& AlexNet & 7/9 & 9/9 & 5/9 & 5/9\\
\hline
ImageNet & ResNet-50 & 4/5 & 4/5 & 4/5 & 1/5\\

& VGG16 & 4/5 & 3/5 & 3/5 & 2/5\\

& AlexNet & 4/5 & 4/5 & 4/5 & 1/5\\
\hline
Total &  & \textbf{67}/85 & \textbf{74}/85 & 54/85 & 48/85 \\
\hline
\hline
\end{tabular}}
\end{center}
\end{table}

\noindent\textbf{Varying number of data samples in each class.}
We also vary the number of validation data points for CIFAR-10 models and see the detection performance when we choose the quantile to be $0.5$ (median). The number of data points in each class is chosen as $1$, $2$, and $5$ and the corresponding AUC values are $0.96, 0.98$ and $1$, respectively. We can see that the data-limited TrojanNet detector is effective even when only one data point is available for each class.

\begin{figure}[h]
\begin{center}
   \includegraphics[width=0.6\linewidth]{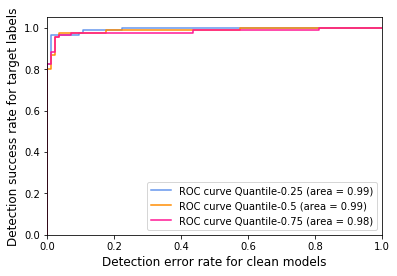}
\end{center}
   \caption{ROC curve for target label detection using data-limited TrojanNet detector over $85$ TrojanNets and $85$ cleanNets}
\label{fig:target_label_d}
\end{figure}

\noindent\textbf{ROC curve for target label detection.}
Let the true positive rate be the  detection success rate of target labels and the false negative rate be the  detection error rate of cleanNets. Table \ref{model_detect} also shows the AUC values for target label detection, and Fig.~\ref{fig:target_label_d} shows the ROC curves for target label detection using DL-TND. We set $I^{(k)}$ to quantile-$0.25$, median, quantile-$0.75$ of the similarity values and vary $T_1$. Under the three different quantile selections, AUC values are all above $0.98$.

\begin{figure}[h]
  \centering
  \begin{adjustbox}{max width=1\textwidth }
    \begin{tabular}{@{\hskip 0.00in}c  @{\hskip 0.02in} c @{\hskip 0.02in} @{\hskip 0.02in} c @{\hskip 0.02in} @{\hskip 0.02in} c}
  & 
\begin{tabular}{@{\hskip 0.00in}  @{\hskip 0.02in}c}
\centering\colorbox{lightgray}{\large \textbf{CIFAR-10 input}}
\end{tabular}

&
\begin{tabular}{@{\hskip 0.00in}  @{\hskip 0.02in}c}
\centering\colorbox{lightgray}{\large \textbf{GTSRB input}}
\end{tabular}

&
\begin{tabular}{@{\hskip 0.00in}  @{\hskip 0.02in}c}
\centering\colorbox{lightgray}{\large \textbf{ImageNet input}}
\end{tabular}

\\

 \begin{tabular}{@{}c@{}}  
\vspace*{0.01in}\\
\rotatebox{90}{\parbox{9em}{\centering \normalsize \textbf{Seed Images}}}
 \\
\rotatebox{90}{\parbox{9em}{\centering \normalsize \textbf{\begin{tabular}[c]{@{}c@{}}Recovered\\ images\\(cleanNet)  \end{tabular}  }}}
 \\
\rotatebox{90}{\parbox{9em}{\centering \normalsize \textbf{\begin{tabular}[c]{@{}c@{}}Perturbation\\ pattern\\(cleanNet)  \end{tabular}}}}
\\
\rotatebox{90}{\parbox{9em}{\centering \normalsize \textbf{Trojan triggers}}}
 \\
\rotatebox{90}{\parbox{9em}{\centering \normalsize \textbf{\begin{tabular}[c]{@{}c@{}}Recovered\\ images\\(TrojanNet)  \end{tabular} }}}
\\
\rotatebox{90}{\parbox{9em}{\centering \normalsize \textbf{\begin{tabular}[c]{@{}c@{}}Perturbation\\ pattern\\(TrojanNet)  \end{tabular} }}}
\\
\end{tabular} 
&
 \begin{tabular}{@{\hskip 0.02in}c@{\hskip 0.02in}}
 \begin{tabular}{@{\hskip 0.02in}c@{\hskip 0.02in}c@{\hskip 0.02in}}
\begin{tabular}{@{\hskip 0.02in}c@{\hskip 0.02in}}
\\
 \parbox[c]{9em}{\includegraphics[width=9em]{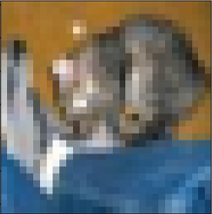}} 
 \\
 \parbox[c]{9em}{\includegraphics[width=9em]{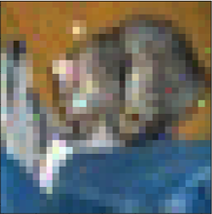}}
  \\
   \parbox[c]{9em}{\includegraphics[width=9em]{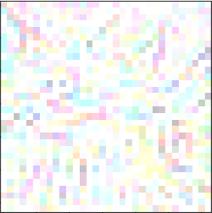}}
  \\
     \parbox[c]{9em}{\includegraphics[width=9em]{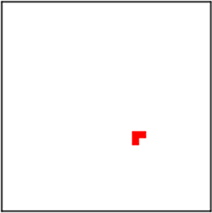}}
  \\
   \parbox[c]{9em}{\includegraphics[width=9em]{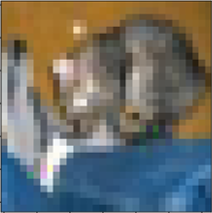}}
  \\
 \parbox[c]{9em}{\includegraphics[width=9em]{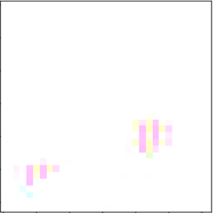}}
 \\
\end{tabular}

&
 \begin{tabular}{@{\hskip 0.02in}c@{\hskip 0.02in}}
\\
 \parbox[c]{9em}{\includegraphics[width=9em]{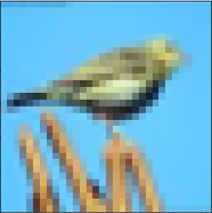}} 
 \\
 \parbox[c]{9em}{\includegraphics[width=9em]{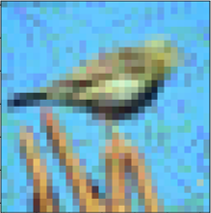}}
  \\
   \parbox[c]{9em}{\includegraphics[width=9em]{Fig/cifar_dl_clean_pert1}}
  \\
     \parbox[c]{9em}{\includegraphics[width=9em]{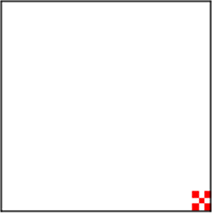}}
  \\
   \parbox[c]{9em}{\includegraphics[width=9em]{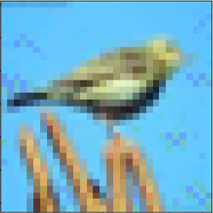}}
  \\
 \parbox[c]{9em}{\includegraphics[width=9em]{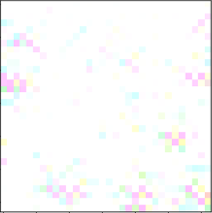}}
 \\
\end{tabular}

\end{tabular}
\end{tabular}
&
 \begin{tabular}{@{\hskip 0.02in}c@{\hskip 0.02in}}
 \begin{tabular}{@{\hskip 0.02in}c@{\hskip 0.02in}c@{\hskip 0.02in}}
\begin{tabular}{@{\hskip 0.02in}c@{\hskip 0.02in}}
\\
 \parbox[c]{9em}{\includegraphics[width=9em]{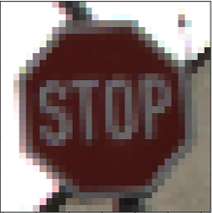}} 
 \\
 \parbox[c]{9em}{\includegraphics[width=9em]{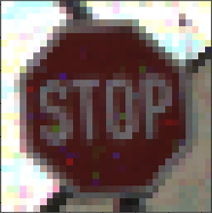}}
  \\
   \parbox[c]{9em}{\includegraphics[width=9em]{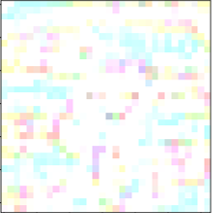}}
  \\
     \parbox[c]{9em}{\includegraphics[width=9em]{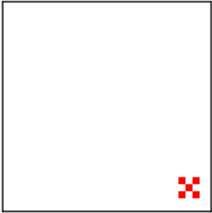}}
  \\
   \parbox[c]{9em}{\includegraphics[width=9em]{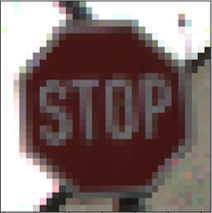}}
  \\
 \parbox[c]{9em}{\includegraphics[width=9em]{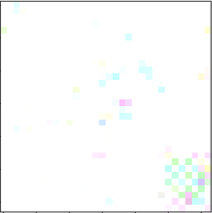}}
 \\
\end{tabular}

&
 \begin{tabular}{@{\hskip 0.02in}c@{\hskip 0.02in}}
\\
 \parbox[c]{9em}{\includegraphics[width=9em]{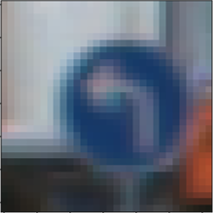}} 
 \\
 \parbox[c]{9em}{\includegraphics[width=9em]{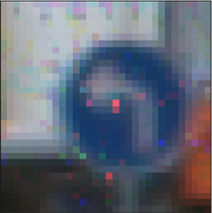}}
  \\
   \parbox[c]{9em}{\includegraphics[width=9em]{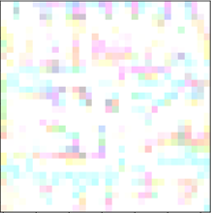}}
  \\
     \parbox[c]{9em}{\includegraphics[width=9em]{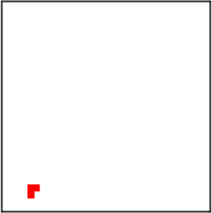}}
  \\
   \parbox[c]{9em}{\includegraphics[width=9em]{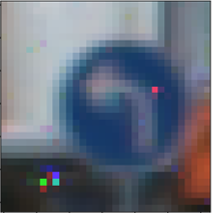}}
  \\
 \parbox[c]{9em}{\includegraphics[width=9em]{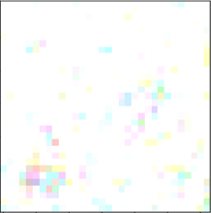}}
 \\

\end{tabular}

\end{tabular}
\end{tabular}
&
 \begin{tabular}{@{\hskip 0.02in}c@{\hskip 0.02in}}
 \begin{tabular}{@{\hskip 0.02in}c @{\hskip 0.02in}c@{\hskip 0.02in}}
\begin{tabular}{@{\hskip 0.02in}c@{\hskip 0.02in}}
\\
 \parbox[c]{9em}{\includegraphics[width=9em]{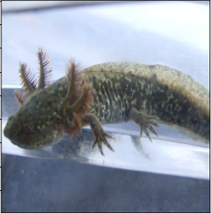}} 
 \\
 \parbox[c]{9em}{\includegraphics[width=9em]{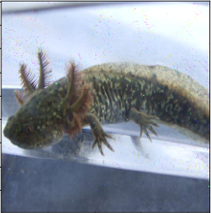}}
  \\
   \parbox[c]{9em}{\includegraphics[width=9em]{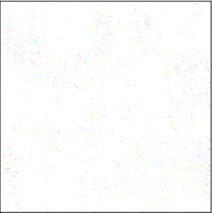}}
  \\
     \parbox[c]{9em}{\includegraphics[width=9em]{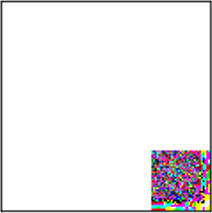}}
  \\
   \parbox[c]{9em}{\includegraphics[width=9em]{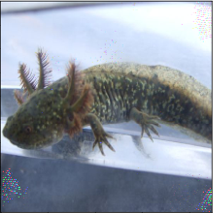}}
  \\
 \parbox[c]{9em}{\includegraphics[width=9em]{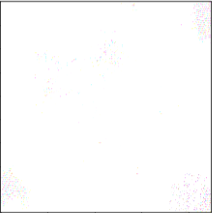}}
 \\
\end{tabular}

&
 \begin{tabular}{@{\hskip 0.02in}c@{\hskip 0.02in}}
\\
 \parbox[c]{9em}{\includegraphics[width=9em]{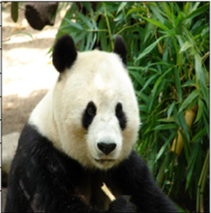}} 
 \\
 \parbox[c]{9em}{\includegraphics[width=9em]{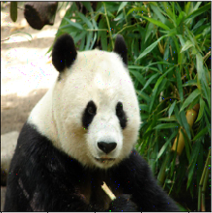}}
  \\
   \parbox[c]{9em}{\includegraphics[width=9em]{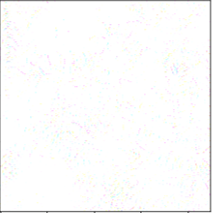}}
  \\
     \parbox[c]{9em}{\includegraphics[width=9em]{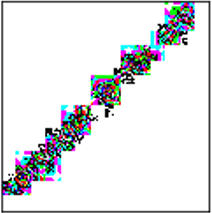}}
  \\
   \parbox[c]{9em}{\includegraphics[width=9em]{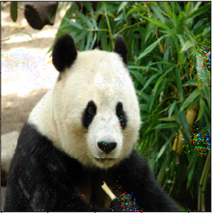}}
  \\
 \parbox[c]{9em}{\includegraphics[width=9em]{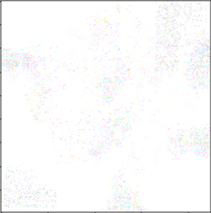}}
 \\

\end{tabular}

\end{tabular}
\end{tabular}

\vspace*{-0.001in}
\end{tabular}
  \end{adjustbox}
    \caption{\small{Visualization of the universal perturbations obtained by our proposed DL-TND. Here the TrojanNets are trained by 10\% poisoned data and clean data, respectively. First row: Seed input images (from left to right: $2$ randomly selected CIFAR-10 images, $2$ randomly selected GTSRB images, $2$ randomly selected ImageNet images). 
Second row: Recovered images under cleanNets. 
Third row: Perturbation patterns given by the difference between the recovered images in the second row and the original seed image. Fourth row: Trojan triggers used for TrojanNets. Fifth row: Recovered images under TrojanNets.
Sixth row: Perturbation patterns given by the difference between the recovered images in the fifth row and the original seed images.  Perturbation has common patterns with the real Trojan triggers. Since DL-TND tries to find the smallest universal perturbation, the recovered perturbation pattern could be much more sparse than the Trojan trigger when the Trojan trigger is very complicated. This can be viewed in the perturbation pattern in the last two columns.
   }}
  \label{fig:rec_trig_DL}
 \vspace*{-1mm}
\end{figure}

\noindent\textbf{Visualization of the universal perturbations.}
In Fig.~\ref{fig:rec_trig_DL}, we show the universal perturbation obtained through \eqref{eq: prob_univ}. Due to the presence of backdoor in TrojanNets, universal perturbations can reveal \textit{common patterns} with the real triggers, and this property is reflected in Fig.~\ref{fig:rec_trig_DL}. Since DL-TND tries to find the smallest universal perturbation, the recovered perturbation pattern could be much more sparse than the Trojan trigger when the Trojan trigger is very complicated. This can be viewed in the perturbation pattern in the last two columns of Fig.~\ref{fig:rec_trig_DL}.

\section{DF-TND: Additional Experiments}
\begin{figure}[h]
  \centering
  \begin{adjustbox}{max width=1\textwidth }
  \begin{tabular}{@{\hskip 0.00in}c  @{\hskip 0.02in}c @{\hskip 0.02in} c @{\hskip 0.02in}c  @{\hskip 0.02in} c @{\hskip 0.02in}c @{\hskip 0.02in} c @{\hskip 0.02in}c @{\hskip 0.02in}c @{\hskip 0.02in}c}
 \begin{tabular}{@{}c@{}}  
\vspace*{0.1in}\\
\rotatebox{90}{\parbox{10em}{\centering \small \textbf{Trojan triggers}}}
 \\
\rotatebox{90}{\parbox{10em}{\centering \small \textbf{Clean Input 1}}}
 \\
\rotatebox{90}{\parbox{10em}{\centering \small \textbf{Perturbations 1}}}
 \\
 \rotatebox{90}{\parbox{10em}{\centering \small \textbf{Clean Input 2}}}
 \\
\rotatebox{90}{\parbox{10em}{\centering \small \textbf{Perturbations 2}}}
 \\
\rotatebox{90}{\parbox{10em}{\centering \small \textbf{Noise Input 3}}}
 \\
\rotatebox{90}{\parbox{10em}{\centering \small \textbf{Perturbations 3}}}
 \\
 \rotatebox{90}{\parbox{10em}{\centering \small \textbf{Noise Input 4}}}
 \\
\rotatebox{90}{\parbox{10em}{\centering \small \textbf{Perturbations 4}}}
\end{tabular} 
&
\begin{tabular}{@{\hskip 0.02in}c@{\hskip 0.02in}}
     \begin{tabular}{@{\hskip 0.00in}c@{\hskip 0.00in}}
     \parbox{10em}{\centering \small  center: [5,5]}  
    \end{tabular} 
    \\
 \parbox[c]{10em}{\includegraphics[width=10em]{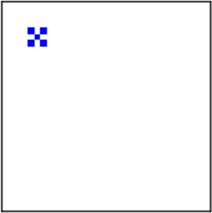}} 
 \\
 \parbox[c]{10em}{\includegraphics[width=10em]{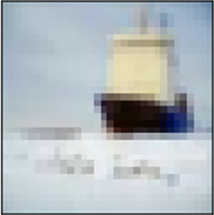}}
  \\
 \parbox[c]{10em}{\includegraphics[width=10em]{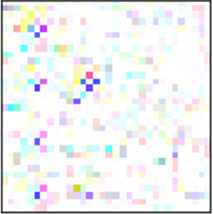}}
  \\
 \parbox[c]{10em}{\includegraphics[width=10em]{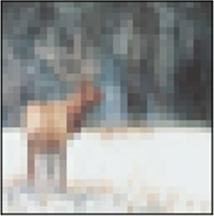}}
     \\
 \parbox[c]{10em}{\includegraphics[width=10em]{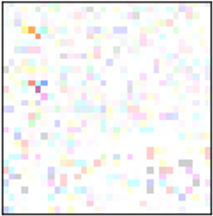}} 
 \\
 \parbox[c]{10em}{\includegraphics[width=10em]{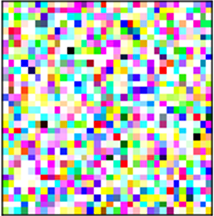}}
  \\
 \parbox[c]{10em}{\includegraphics[width=10em]{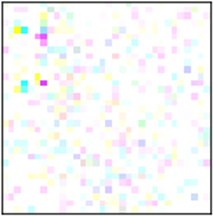}}
  \\
 \parbox[c]{10em}{\includegraphics[width=10em]{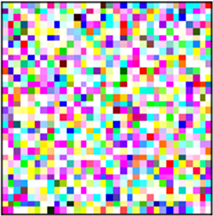}}
   \\
 \parbox[c]{10em}{\includegraphics[width=10em]{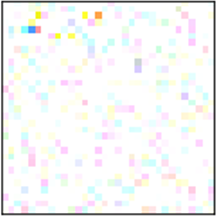}}
\end{tabular}
&
 \begin{tabular}{@{\hskip 0.02in}c@{\hskip 0.02in}}
      \begin{tabular}{@{\hskip 0.00in}c@{\hskip 0.00in}}
     \parbox{10em}{\centering \small center: [5,15]}
    \end{tabular} 
    \\
 \parbox[c]{10em}{\includegraphics[width=10em]{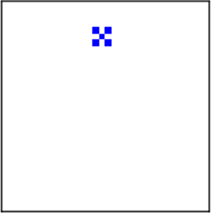}} 
 \\
 \parbox[c]{10em}{\includegraphics[width=10em]{Fig/fl11.png}}
  \\
 \parbox[c]{10em}{\includegraphics[width=10em]{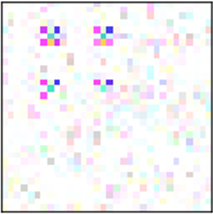}}
  \\
 \parbox[c]{10em}{\includegraphics[width=10em]{Fig/fl12}}
     \\
 \parbox[c]{10em}{\includegraphics[width=10em]{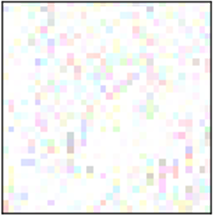}} 
 \\
 \parbox[c]{10em}{\includegraphics[width=10em]{Fig/nl11.png}}
  \\
 \parbox[c]{10em}{\includegraphics[width=10em]{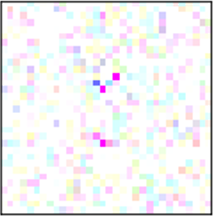}}
  \\
 \parbox[c]{10em}{\includegraphics[width=10em]{Fig/nl12}}
  \\
 \parbox[c]{10em}{\includegraphics[width=10em]{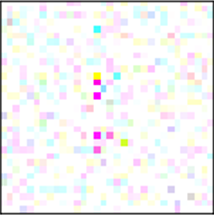}}
\end{tabular}
&
 \begin{tabular}{@{\hskip 0.02in}c@{\hskip 0.02in}}
      \begin{tabular}{@{\hskip 0.00in}c@{\hskip 0.00in}}
     \parbox{10em}{\centering \small  center: [5,25]} 
    \end{tabular} 
    \\
 \parbox[c]{10em}{\includegraphics[width=10em]{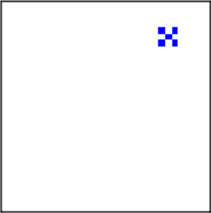}} 
 \\
 \parbox[c]{10em}{\includegraphics[width=10em]{Fig/fl11.png}} 
  \\
 \parbox[c]{10em}{\includegraphics[width=10em]{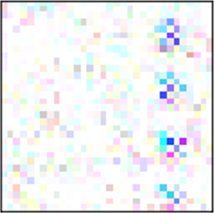}}
  \\
 \parbox[c]{10em}{\includegraphics[width=10em]{Fig/fl12}}
    \\
 \parbox[c]{10em}{\includegraphics[width=10em]{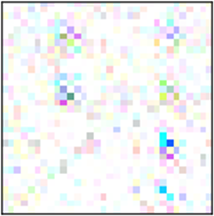}} 
 \\
 \parbox[c]{10em}{\includegraphics[width=10em]{Fig/nl11.png}} 
  \\
 \parbox[c]{10em}{\includegraphics[width=10em]{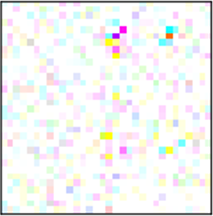}}
  \\
 \parbox[c]{10em}{\includegraphics[width=10em]{Fig/nl12}}
  \\
 \parbox[c]{10em}{\includegraphics[width=10em]{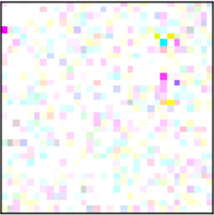}}
\end{tabular}
&
 \begin{tabular}{@{\hskip 0.02in}c@{\hskip 0.02in}c@{\hskip 0.02in} }
      \begin{tabular}{@{\hskip 0.00in}c@{\hskip 0.00in}}
     \parbox{10em}{\centering \small  center: [15,5]}   
    \end{tabular} 
   \\
 \parbox[c]{10em}{\includegraphics[width=10em]{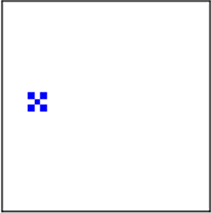}} 
 \\
 \parbox[c]{10em}{\includegraphics[width=10em]{Fig/fl11.png}} 
 \\
 \parbox[c]{10em}{\includegraphics[width=10em]{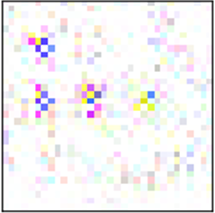}}
 \\
 \parbox[c]{10em}{\includegraphics[width=10em]{Fig/fl12}}
   \\
 \parbox[c]{10em}{\includegraphics[width=10em]{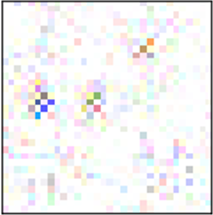}} 
 \\
 \parbox[c]{10em}{\includegraphics[width=10em]{Fig/nl11.png}} 
 \\
 \parbox[c]{10em}{\includegraphics[width=10em]{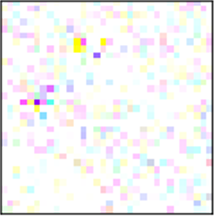}}
 \\
 \parbox[c]{10em}{\includegraphics[width=10em]{Fig/nl12}}
 \\
 \parbox[c]{10em}{\includegraphics[width=10em]{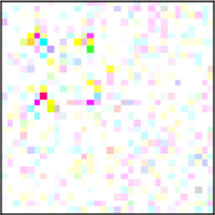}}
\end{tabular}
&
 \begin{tabular}{@{\hskip 0.02in}c@{\hskip 0.02in}}
      \begin{tabular}{@{\hskip 0.00in}c@{\hskip 0.00in}}
     \parbox{10em}{\centering \small  center: [15,15]} 
    \end{tabular} 
    \\
 \parbox[c]{10em}{\includegraphics[width=10em]{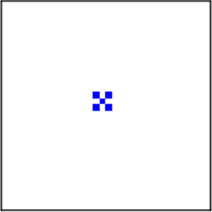}} 
 \\
 \parbox[c]{10em}{\includegraphics[width=10em]{Fig/fl11.png}} 
  \\
 \parbox[c]{10em}{\includegraphics[width=10em]{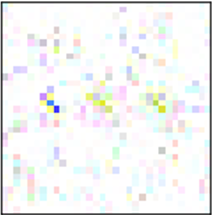}}
  \\
 \parbox[c]{10em}{\includegraphics[width=10em]{Fig/fl12}}
     \\
 \parbox[c]{10em}{\includegraphics[width=10em]{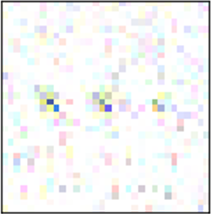}} 
 \\
 \parbox[c]{10em}{\includegraphics[width=10em]{Fig/nl11.png}} 
  \\
 \parbox[c]{10em}{\includegraphics[width=10em]{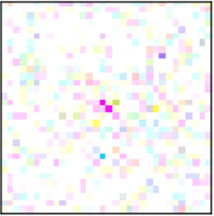}}
  \\
 \parbox[c]{10em}{\includegraphics[width=10em]{Fig/nl12}}
  \\
 \parbox[c]{10em}{\includegraphics[width=10em]{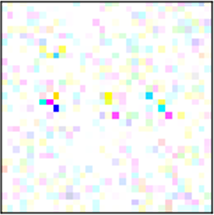}}
\end{tabular}
&
 \begin{tabular}{@{\hskip 0.02in}c@{\hskip 0.02in}}
      \begin{tabular}{@{\hskip 0.00in}c@{\hskip 0.00in}}
     \parbox{10em}{\centering \small  center: [15,25]} 
    \end{tabular} 
    \\
 \parbox[c]{10em}{\includegraphics[width=10em]{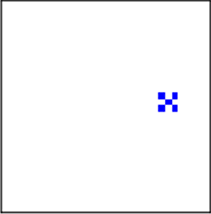}} 
 \\
 \parbox[c]{10em}{\includegraphics[width=10em]{Fig/fl11}}
 \\
 \parbox[c]{10em}{\includegraphics[width=10em]{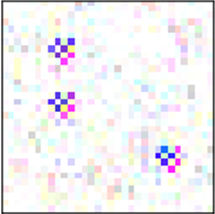}}
 \\
 \parbox[c]{10em}{\includegraphics[width=10em]{Fig/fl12}}
    \\
 \parbox[c]{10em}{\includegraphics[width=10em]{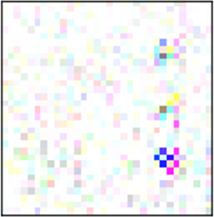}} 
 \\
 \parbox[c]{10em}{\includegraphics[width=10em]{Fig/nl11}}
 \\
 \parbox[c]{10em}{\includegraphics[width=10em]{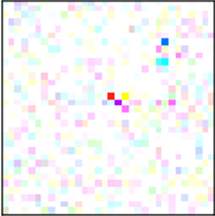}}
 \\
 \parbox[c]{10em}{\includegraphics[width=10em]{Fig/nl12}}
 \\
 \parbox[c]{10em}{\includegraphics[width=10em]{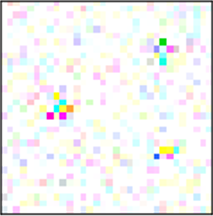}}
\end{tabular}
&
 \begin{tabular}{@{\hskip 0.02in}c@{\hskip 0.02in}}
      \begin{tabular}{@{\hskip 0.00in}c@{\hskip 0.00in}}
     \parbox{10em}{\centering \small  center: [25,5]} 
    \end{tabular} 
    \\
 \parbox[c]{10em}{\includegraphics[width=10em]{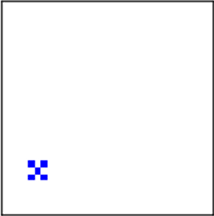}} 
 \\
 \parbox[c]{10em}{\includegraphics[width=10em]{Fig/fl11}} 
 \\
 \parbox[c]{10em}{\includegraphics[width=10em]{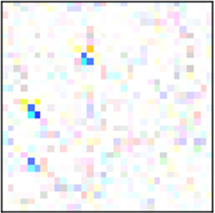}}
 \\
 \parbox[c]{10em}{\includegraphics[width=10em]{Fig/fl12}}
    \\
 \parbox[c]{10em}{\includegraphics[width=10em]{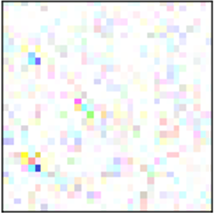}} 
 \\
 \parbox[c]{10em}{\includegraphics[width=10em]{Fig/nl11}} 
 \\
 \parbox[c]{10em}{\includegraphics[width=10em]{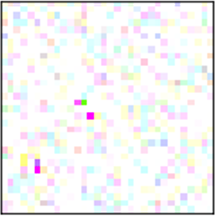}}
 \\
 \parbox[c]{10em}{\includegraphics[width=10em]{Fig/nl12}}
  \\
 \parbox[c]{10em}{\includegraphics[width=10em]{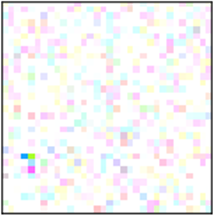}}
\end{tabular}
&
 \begin{tabular}{@{\hskip 0.02in}c@{\hskip 0.02in}}
      \begin{tabular}{@{\hskip 0.00in}c@{\hskip 0.00in}}
     \parbox{10em}{\centering \small  center: [25,15]} 
    \end{tabular} 
    \\
 \parbox[c]{10em}{\includegraphics[width=10em]{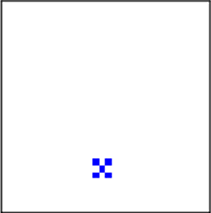}} 
 \\
 \parbox[c]{10em}{\includegraphics[width=10em]{Fig/fl11}} 
  \\
 \parbox[c]{10em}{\includegraphics[width=10em]{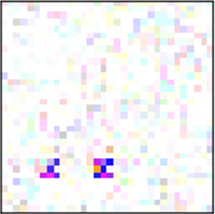}}
  \\
 \parbox[c]{10em}{\includegraphics[width=10em]{Fig/fl12}}
     \\
 \parbox[c]{10em}{\includegraphics[width=10em]{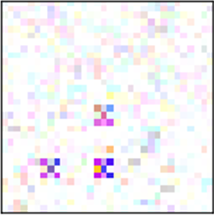}} 
 \\
 \parbox[c]{10em}{\includegraphics[width=10em]{Fig/nl11}} 
  \\
 \parbox[c]{10em}{\includegraphics[width=10em]{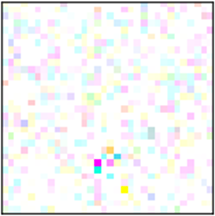}}
  \\
 \parbox[c]{10em}{\includegraphics[width=10em]{Fig/nl12}}
  \\
 \parbox[c]{10em}{\includegraphics[width=10em]{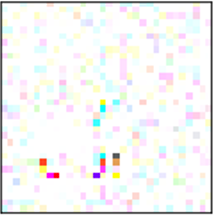}}
\end{tabular}
&
 \begin{tabular}{@{\hskip 0.02in}c@{\hskip 0.02in}}
      \begin{tabular}{@{\hskip 0.00in}c@{\hskip 0.00in}}
     \parbox{10em}{\centering \small  center: [25,25]} 
    \end{tabular} 
    \\
 \parbox[c]{10em}{\includegraphics[width=10em]{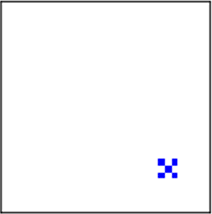}} 
 \\
 \parbox[c]{10em}{\includegraphics[width=10em]{Fig/fl11}} 
  \\
 \parbox[c]{10em}{\includegraphics[width=10em]{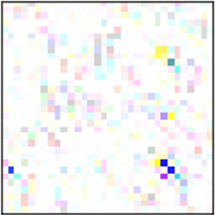}}
  \\
 \parbox[c]{10em}{\includegraphics[width=10em]{Fig/fl12}}
    \\
 \parbox[c]{10em}{\includegraphics[width=10em]{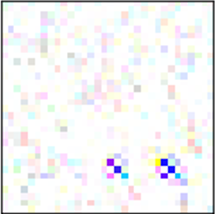}} 
 \\
 \parbox[c]{10em}{\includegraphics[width=10em]{Fig/nl11}} 
  \\
 \parbox[c]{10em}{\includegraphics[width=10em]{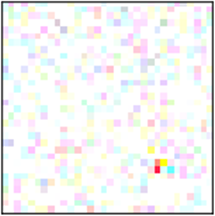}}
  \\
 \parbox[c]{10em}{\includegraphics[width=10em]{Fig/nl12}}
   \\
 \parbox[c]{10em}{\includegraphics[width=10em]{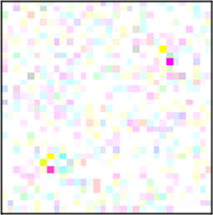}}
\end{tabular}

\vspace*{-0.001in}
\end{tabular}
  \end{adjustbox}
    \caption{\small{Visualization of perturbations when locations of Trojan triggers change. Here the TrojanNets are trained by 10\% poisoned data and clean data, respectively. First row: Trojan triggers in different locations (the centers of the triggers are listed above).
Second row: Clean seed image 1 (ship). 
Third row: Recovered perturbations 1 with input from the second row. Fourth row: Clean seed image 2 (deer). 
Fifth row: Recovered perturbations 2 with input from the fourth row. Sixth row: random noise seed image 3. 
Seventh row: Recovered perturbations 3 with input from the sixth row. Eighth row: random noise seed image 4. 
Ninth row: Recovered perturbations 4 with input from the eighth row. Locations of perturbations vary when the locations of Trojan triggers vary. However, the recovered perturbations do not always have the same locations as the Trojan triggers. Patterns shifted and enlarged due to the convolution operations.
   }}
  \label{fig: vis_loc_vary}
 \vspace*{-1mm}
\end{figure}

\begin{figure}[h]
  \centering
  \begin{adjustbox}{max width=1\textwidth }
    \begin{tabular}{@{\hskip 0.00in}c  @{\hskip 0.02in} c @{\hskip 0.02in} c @{\hskip 0.02in} c @{\hskip 0.02in} c }
  & 
  \begin{tabular}{@{\hskip 0.00in}c  @{\hskip 0.02in} c @{\hskip 0.02in} @{\hskip 0.02in} c }
\centering\colorbox{lightgray}{\large \textbf{trigger size $3\times 3$}}
&
\hspace*{0.8in}
\centering\colorbox{lightgray}{\large \textbf{trigger size $5\times 5$}}
\end{tabular}

&
  \begin{tabular}{@{\hskip 0.00in}c  @{\hskip 0.02in} c @{\hskip 0.02in} @{\hskip 0.02in} c }
\centering\colorbox{lightgray}{\large \textbf{trigger size $7\times 7$}}
&\hspace*{0.8in}
\centering\colorbox{lightgray}{\large \textbf{trigger size $9\times 9$}}
\end{tabular}

\\

 \begin{tabular}{@{}c@{}}  
\vspace*{0.01in}\\
\rotatebox{90}{\parbox{9em}{\centering \normalsize \textbf{Trojan triggers}}}
 \\
\rotatebox{90}{\parbox{9em}{\centering \normalsize \textbf{Seed images}}}
 \\
\rotatebox{90}{\parbox{9em}{\centering \normalsize \textbf{\begin{tabular}[c]{@{}c@{}}Recovered\\ images  \end{tabular} }}}
\\
\rotatebox{90}{\parbox{9em}{\centering \normalsize \textbf{\begin{tabular}[c]{@{}c@{}}Perturbation\\ pattern  \end{tabular} }}}

\end{tabular} 
&
 \begin{tabular}{@{\hskip 0.02in}c@{\hskip 0.02in}c@{\hskip 0.02in}}
 \begin{tabular}{@{\hskip 0.02in}c@{\hskip 0.02in}c@{\hskip 0.02in}c@{\hskip 0.02in}c@{\hskip 0.02in}}
\begin{tabular}{@{\hskip 0.02in}c@{\hskip 0.02in}}
\\
 \parbox[c]{9em}{\includegraphics[width=9em]{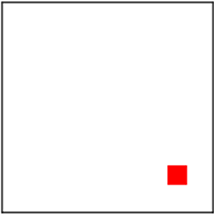}} 
 \\
 \parbox[c]{9em}{\includegraphics[width=9em]{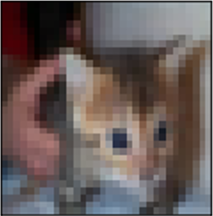}}
  \\
   \parbox[c]{9em}{\includegraphics[width=9em]{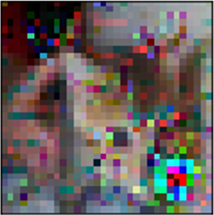}}
  \\
   \parbox[c]{9em}{\includegraphics[width=9em]{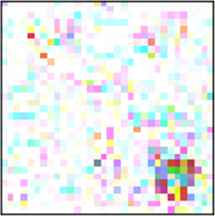}}

\end{tabular}

&
 \begin{tabular}{@{\hskip 0.02in}c@{\hskip 0.02in}}
\\
 \parbox[c]{9em}{\includegraphics[width=9em]{Fig/trigsize3}} 
 \\
 \parbox[c]{9em}{\includegraphics[width=9em]{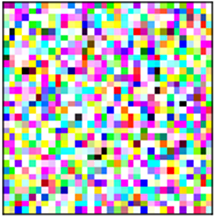}}
  \\
  \parbox[c]{9em}{\includegraphics[width=9em]{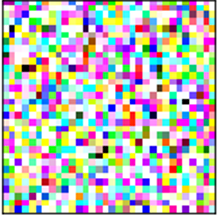}} 
  \\
   \parbox[c]{9em}{\includegraphics[width=9em]{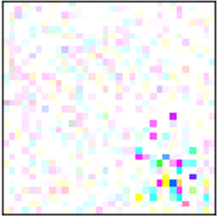}} 

\end{tabular}
&
\begin{tabular}{@{\hskip 0.02in}c@{\hskip 0.02in}c@{\hskip 0.02in}}
\\
 \parbox[c]{9em}{\includegraphics[width=9em]{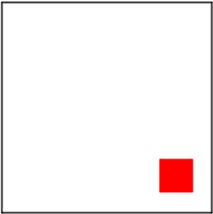}} 
 \\
 \parbox[c]{9em}{\includegraphics[width=9em]{Fig/sizetestclean}}
  \\
   \parbox[c]{9em}{\includegraphics[width=9em]{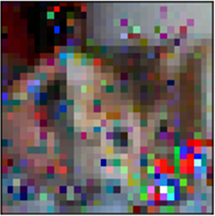}}
  \\
   \parbox[c]{9em}{\includegraphics[width=9em]{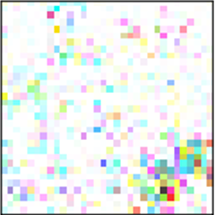}}

\end{tabular}
&
 \begin{tabular}{@{\hskip 0.02in}c@{\hskip 0.02in}}
\\
 \parbox[c]{9em}{\includegraphics[width=9em]{Fig/trigsize5}} 
 \\
 \parbox[c]{9em}{\includegraphics[width=9em]{Fig/sizetestnoise}}
  \\
   \parbox[c]{9em}{\includegraphics[width=9em]{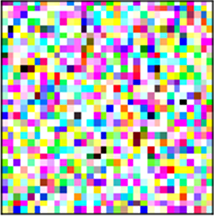}}
  \\
   \parbox[c]{9em}{\includegraphics[width=9em]{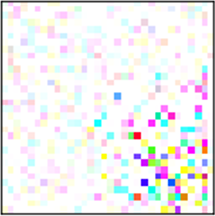}}

\end{tabular}
\end{tabular}
\end{tabular}
&
 \begin{tabular}{@{\hskip 0.02in}c@{\hskip 0.02in}}
 \begin{tabular}{@{\hskip 0.02in}c@{\hskip 0.02in}c@{\hskip 0.02in}c@{\hskip 0.02in}c@{\hskip 0.02in}}
 \begin{tabular}{@{\hskip 0.02in}c@{\hskip 0.02in}}
\\
 \parbox[c]{9em}{\includegraphics[width=9em]{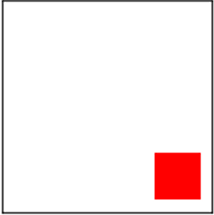}} 
 \\
 \parbox[c]{9em}{\includegraphics[width=9em]{Fig/sizetestclean}} 
  \\
  \parbox[c]{9em}{\includegraphics[width=9em]{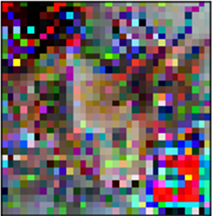}} 
  \\
   \parbox[c]{9em}{\includegraphics[width=9em]{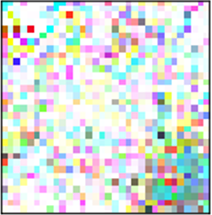}}

\end{tabular}
&
 \begin{tabular}{@{\hskip 0.02in}c@{\hskip 0.02in}}
\\
 \parbox[c]{9em}{\includegraphics[width=9em]{Fig/trigsize7}} 
 \\
 \parbox[c]{9em}{\includegraphics[width=9em]{Fig/sizetestnoise}} 
  \\
  \parbox[c]{9em}{\includegraphics[width=9em]{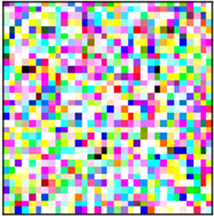}} 
  \\
   \parbox[c]{9em}{\includegraphics[width=9em]{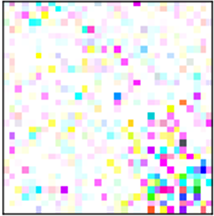}} 

\end{tabular}

&
 \begin{tabular}{@{\hskip 0.02in}c@{\hskip 0.02in}}
\\
 \parbox[c]{9em}{\includegraphics[width=9em]{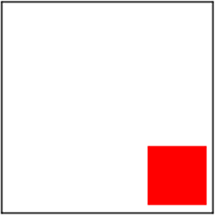}} 
 \\
 \parbox[c]{9em}{\includegraphics[width=9em]{Fig/sizetestclean}}
  \\
   \parbox[c]{9em}{\includegraphics[width=9em]{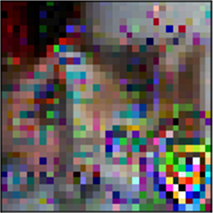}}
  \\
  \parbox[c]{9em}{\includegraphics[width=9em]{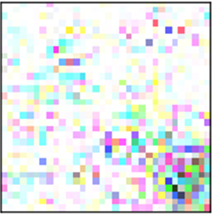}}
 
\end{tabular}
&
\begin{tabular}{@{\hskip 0.02in}c@{\hskip 0.02in}}
\\
 \parbox[c]{9em}{\includegraphics[width=9em]{Fig/trigsize9}} 
 \\
 \parbox[c]{9em}{\includegraphics[width=9em]{Fig/sizetestnoise}}
  \\
   \parbox[c]{9em}{\includegraphics[width=9em]{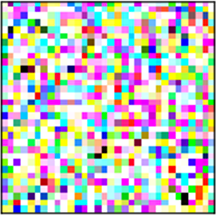}}
  \\
  \parbox[c]{9em}{\includegraphics[width=9em]{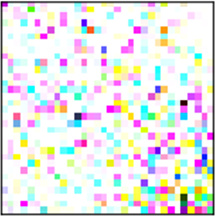}}
 
\end{tabular} 
\end{tabular}
\end{tabular} 

\vspace*{-0.001in}
\end{tabular}
  \end{adjustbox}
    \caption{\small{Visualization of perturbations when sizes of Trojan triggers change. Here the Trojan ResNet-50 models are trained by 10\% poisoned data and clean data, respectively. We vary the trigger size from $3 \times 3$ to $9 \times 9$ and show the recovery in different columns. First row: Trojan triggers with different sizes (the sizes of the triggers are listed above).
Second row: Seed images (clean CIFAR-10 images and random noise images). 
Third row: Recovered images with inputs from the second row. Fourth row: Perturbation patterns given by the difference between the recovered images in the third row and the seed images. One can see that the area of the recovered perturbation increases when the size of the Trojan trigger increases.
   }}
  \label{fig: vis_size_vary}
 \vspace*{-1mm}
\end{figure}

\noindent\textbf{The sensitivity to trigger locations and sizes.}
Fig.~\ref{fig: vis_loc_vary} and Fig.~\ref{fig: vis_size_vary} provide the experimental results for the sensitivity to trigger locations and sizes. Fig.~\ref{fig: vis_loc_vary} shows that locations of perturbations vary when the locations of Trojan triggers vary. However, the recovered perturbations do not always have the same locations as the Trojan triggers. Patterns shifted and enlarged due to the convolution operations. Fig.~\ref{fig: vis_size_vary} shows that DF-TND can recover the trigger pattern when the size of the Trojan trigger increases, and the area of the recovered perturbation increases when the size of the Trojan trigger increases.

\begin{figure}[h]
   \centering
\hspace*{-0.1in}\begin{tabular}{cc}
\includegraphics[width=0.45\columnwidth]{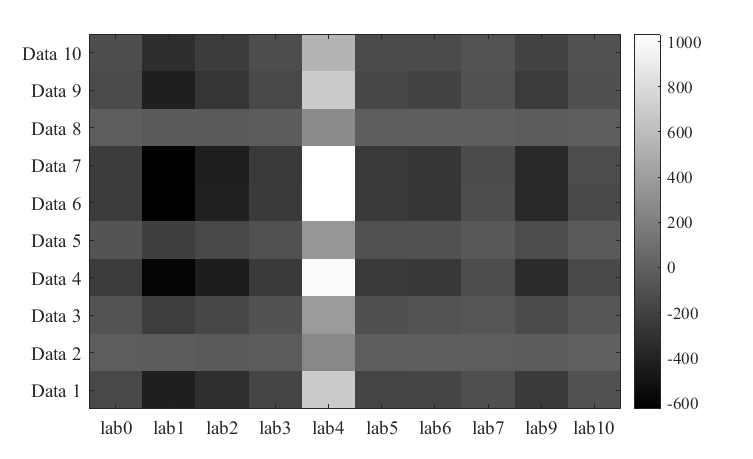}
\vspace*{-0.05in}
&
\includegraphics[width=0.45\columnwidth]{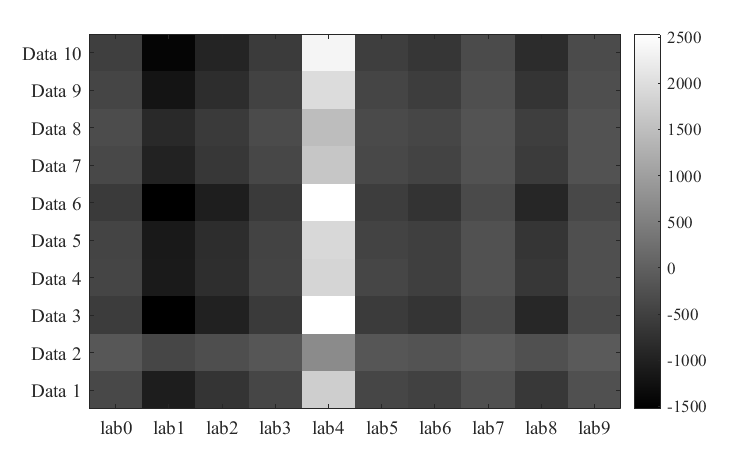}
\\
\scriptsize{(a)} & \scriptsize{(b)}
\end{tabular}
\caption{\small{Improvements using DF-TND together with our refine method - maximizing the neuron activation corresponding to the Trojan-related coordinate: (a) Visualization of logits output increase of 10 random noise inputs before using the refine method when label 4 is the target label. The maximum increase belonging to the target label is $600$, the maximum decreasing belonging to the non-target label is $600$. (b) Visualization of logits output increase of the same 10 random noise inputs after using the refine method when label 4 is the target label. The maximum increase belonging to the target label is $2500$, the maximum decreasing belonging to the non-target label is $1500$. The refine method increases the logits output belonging to the target label, while decreases the logits outputs belonging to the non-target labels
}}
\label{fig: refine}
\end{figure}

\noindent\textbf{Improvements using the refine method - maximizing the neuron activation corresponding to the Trojan-related coordinate.}
Note that once the recovered data is obtained from the optimization problem \eqref{neural_act}, one can find the coordinate related to Trojan feature by checking the largest neuron activation value (or the largest weight) among all the coordinates. Then maximizing the output of the Trojan-related coordinate separately could provide a better result. Fig.~\ref{fig: refine} shows the improvements using DF-TND together with our refine method - maximizing the neuron activation corresponding to the Trojan-related coordinate. The refine method can increase the logits output belonging to the target label, while decrease the logits outputs belonging to the non-target labels simultaneously.

\begin{figure}[h]
  \centering
\hspace*{-0.1in}\begin{tabular}{cc}
\includegraphics[width=0.5\columnwidth]{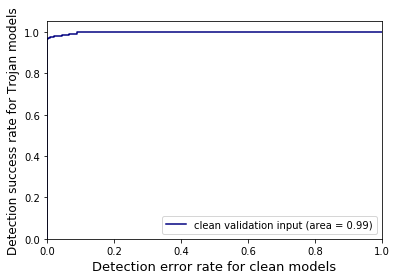}
&
\includegraphics[width=0.5\columnwidth]{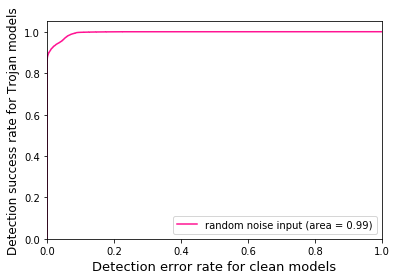}
\vspace*{-0.05in}
\\
(a) & (b)
\end{tabular}
\caption{\small{ROC curves for TrojanNet detection with clean validation inputs and random noise inputs: (a) Clean validation inputs (AUC=$0.99$) (b) Random noise inputs (AUC=$0.99$)
}}
\label{fig: df_mod_detect}
\end{figure}

\noindent\textbf{ROC curves for TrojanNet detection with clean validation inputs and random noise inputs.}
Fig.~\ref{fig: df_mod_detect} (a) and (b) show the ROC curves for TrojanNet detection with clean validation inputs and random noise inputs, respectively. The true positive rate is the detection success rate for TrojanNets and the false negative rate is the detection error rate for cleanNets. In both cases, DF-TND can reach nearly perfect AUC values $0.99$. $T_2=55-400$ could provide a detection success rate of more than  $85\%$ for TrojanNets and a detection success rate of over $90\%$ for cleanNets.

\noindent\textbf{Recovered perturbations under different $\lambda$.}

\begin{figure}[h]
  \centering
  \begin{adjustbox}{max width=1\textwidth }
  \begin{tabular}{@{\hskip 0.00in}c  @{\hskip 0.00in}  c | @{\hskip 0.02in} c @{\hskip 0.02in}c @{\hskip 0.02in} c @{\hskip 0.02in}c @{\hskip 0.02in}c}
 \begin{tabular}{@{}c@{}}  
\vspace*{0.1in}\\
\rotatebox{90}{\parbox{10em}{\centering \small \textbf{TrojanNet}}}
 \\
\rotatebox{90}{\parbox{10em}{\centering \small \textbf{cleanNet}}}
\end{tabular} 
&
\begin{tabular}{@{\hskip 0.02in}c@{\hskip 0.02in}}
     \begin{tabular}{@{\hskip 0.00in}c@{\hskip 0.00in}}
     \parbox{10em}{\centering \small clean inputs}  
    \end{tabular} 
    \\
 \parbox[c]{10em}{\includegraphics[width=10em]{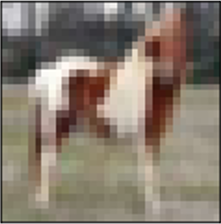}} 
 \\
 \parbox[c]{10em}{\includegraphics[width=10em]{Fig/testfig1.png}}

\end{tabular}
&
 \begin{tabular}{@{\hskip 0.02in}c@{\hskip 0.02in}}
      \begin{tabular}{@{\hskip 0.00in}c@{\hskip 0.00in}}
     \parbox{10em}{\centering \small  Trojan vs. clean}
    \end{tabular} 
    \\
 \parbox[c]{10em}{\includegraphics[width=10em]{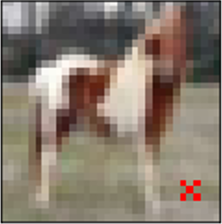}} 
 \\
 \parbox[c]{10em}{\includegraphics[width=10em]{Fig/testfig1}}
\end{tabular}
&
 \begin{tabular}{@{\hskip 0.02in}c@{\hskip 0.02in}}
      \begin{tabular}{@{\hskip 0.00in}c@{\hskip 0.00in}}
     \parbox{10em}{\centering \small  $\lambda = 0.000001$} 
    \end{tabular} 
    \\
 \parbox[c]{10em}{\includegraphics[width=10em]{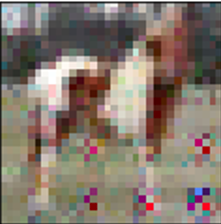}} 
 \\
 \parbox[c]{10em}{\includegraphics[width=10em]{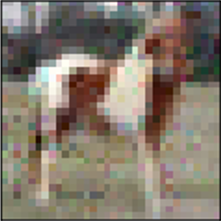}} 
\end{tabular}
&
 \begin{tabular}{@{\hskip 0.02in}c@{\hskip 0.02in}c@{\hskip 0.02in} }
      \begin{tabular}{@{\hskip 0.00in}c@{\hskip 0.00in}}
     \parbox{10em}{\centering \small  $\lambda = 0.00001$}   
    \end{tabular} 
  \\
 \parbox[c]{10em}{\includegraphics[width=10em]{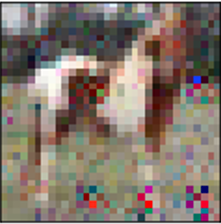}} 
 \\
 \parbox[c]{10em}{\includegraphics[width=10em]{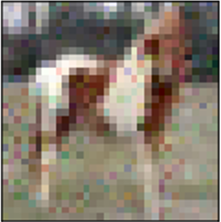}} 
\end{tabular}
&
 \begin{tabular}{@{\hskip 0.02in}c@{\hskip 0.02in}}
      \begin{tabular}{@{\hskip 0.00in}c@{\hskip 0.00in}}
     \parbox{10em}{\centering \small  $\lambda = 0.0001$} 
    \end{tabular} 
    \\
 \parbox[c]{10em}{\includegraphics[width=10em]{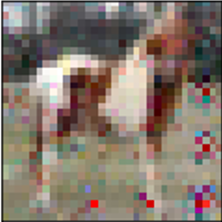}} 
 \\
 \parbox[c]{10em}{\includegraphics[width=10em]{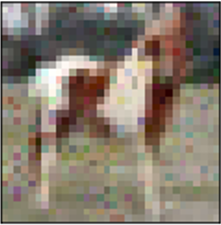}} 
\end{tabular}
&
 \begin{tabular}{@{\hskip 0.02in}c@{\hskip 0.02in}}
      \begin{tabular}{@{\hskip 0.00in}c@{\hskip 0.00in}}
     \parbox{10em}{\centering \small  $\lambda = 0.001$} 
    \end{tabular} 
    \\
 \parbox[c]{10em}{\includegraphics[width=10em]{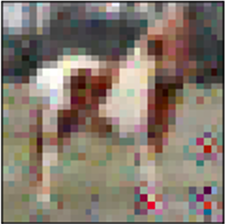}} 
 \\
 \parbox[c]{10em}{\includegraphics[width=10em]{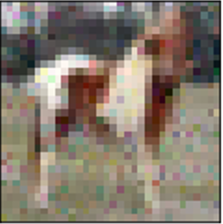}}
\end{tabular}
\vspace*{-0.001in}
\end{tabular}
  \end{adjustbox}
    \caption{\small{Visualization of perturbations with CIFAR-10 image inputs when the sparsity penalty parameter $\lambda$ varies. First row: Images under a TrojanNet. Second row: Images under a cleanNet.
  }}
  \label{fig:cifar_input}
 \vspace*{-1mm}
\end{figure}

\begin{figure}[h]
  \centering
  \begin{adjustbox}{max width=1\textwidth }
  \begin{tabular}{@{\hskip 0.00in}c  @{\hskip 0.00in}  c | @{\hskip 0.02in} c @{\hskip 0.02in}c @{\hskip 0.02in} c @{\hskip 0.02in}c @{\hskip 0.02in}c}
 \begin{tabular}{@{}c@{}}  
\vspace*{0.1in}\\
\rotatebox{90}{\parbox{10em}{\centering \small \textbf{TrojanNet}}}
 \\
\rotatebox{90}{\parbox{10em}{\centering \small \textbf{cleanNet}}}
\end{tabular} 
&
\begin{tabular}{@{\hskip 0.02in}c@{\hskip 0.02in}}
     \begin{tabular}{@{\hskip 0.00in}c@{\hskip 0.00in}}
     \parbox{10em}{\centering \small random noise inputs}  
    \end{tabular} 
    \\
 \parbox[c]{10em}{\includegraphics[width=10em]{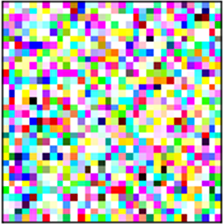}} 
 \\
 \parbox[c]{10em}{\includegraphics[width=10em]{Fig/testnoise1.png}}

\end{tabular}
&
 \begin{tabular}{@{\hskip 0.02in}c@{\hskip 0.02in}}
      \begin{tabular}{@{\hskip 0.00in}c@{\hskip 0.00in}}
     \parbox{10em}{\centering \small  Trojan vs. clean}
    \end{tabular} 
    \\
 \parbox[c]{10em}{\includegraphics[width=10em]{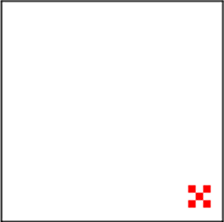}} 
 \\
 \parbox[c]{10em}{\includegraphics[width=10em]{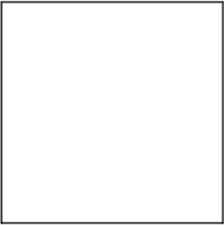}}
\end{tabular}
&
 \begin{tabular}{@{\hskip 0.02in}c@{\hskip 0.02in}}
      \begin{tabular}{@{\hskip 0.00in}c@{\hskip 0.00in}}
     \parbox{10em}{\centering \small  $\lambda = 0.000001$} 
    \end{tabular} 
    \\
 \parbox[c]{10em}{\includegraphics[width=10em]{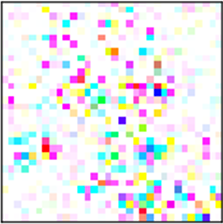}} 
 \\
 \parbox[c]{10em}{\includegraphics[width=10em]{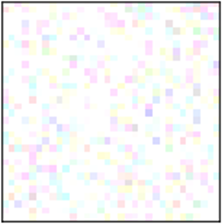}} 
\end{tabular}
&
 \begin{tabular}{@{\hskip 0.02in}c@{\hskip 0.02in}c@{\hskip 0.02in} }
      \begin{tabular}{@{\hskip 0.00in}c@{\hskip 0.00in}}
     \parbox{10em}{\centering \small  $\lambda = 0.00001$}   
    \end{tabular} 
  \\
 \parbox[c]{10em}{\includegraphics[width=10em]{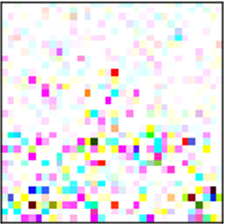}} 
 \\
 \parbox[c]{10em}{\includegraphics[width=10em]{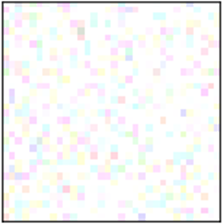}} 
\end{tabular}
&
 \begin{tabular}{@{\hskip 0.02in}c@{\hskip 0.02in}}
      \begin{tabular}{@{\hskip 0.00in}c@{\hskip 0.00in}}
     \parbox{10em}{\centering \small  $\lambda = 0.0001$} 
    \end{tabular} 
    \\
 \parbox[c]{10em}{\includegraphics[width=10em]{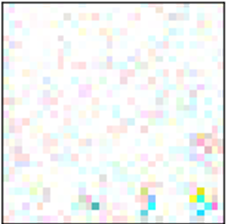}} 
 \\
 \parbox[c]{10em}{\includegraphics[width=10em]{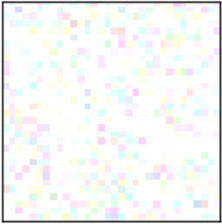}} 
\end{tabular}
&
 \begin{tabular}{@{\hskip 0.02in}c@{\hskip 0.02in}}
      \begin{tabular}{@{\hskip 0.00in}c@{\hskip 0.00in}}
     \parbox{10em}{\centering \small  $\lambda = 0.001$} 
    \end{tabular} 
    \\
 \parbox[c]{10em}{\includegraphics[width=10em]{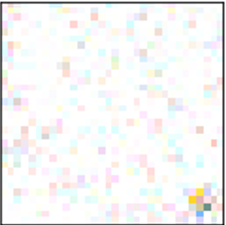}} 
 \\
 \parbox[c]{10em}{\includegraphics[width=10em]{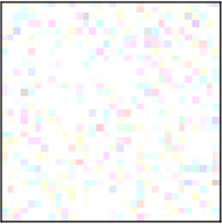}}
\end{tabular}
\vspace*{-0.001in}
\end{tabular}
  \end{adjustbox}
    \caption{\small{Visualization of perturbations with random noise inputs when the sparsity penalty parameter $\lambda$ varies. First row: Images under a TrojanNet. Second row: Images under a cleanNet.
  }}
  \label{fig:noise_input}
 \vspace*{-1mm}
\end{figure}

In Fig.~\ref{fig:cifar_input} and Fig.~\ref{fig:noise_input}, we vary the sparsity penalty parameter $\lambda$ and obtain perturbations under a TrojanNet and a cleanNet. One can find that the trigger pattern appears in the perturbations under the TrojanNet. The perturbations under cleanNet behave like random noises. Another discovery is that the perturbations become more and more sparse when $\lambda$ increases.

This method also works for random noise inputs. Fig.~\ref{fig:noise_input} shows the original noise images, trigger, perturbations under poisoned model, and perturbations under cleanNet. The same behaviors as the clean inputs are observed.

\end{document}